\documentclass[10pt,journal,compsoc]{IEEEtran}
\ifCLASSOPTIONcompsoc
\usepackage[nocompress]{cite}
\else
\usepackage{cite}
\fi

\usepackage{setspace}
\usepackage{graphicx}
\usepackage{amsmath,amsfonts}
\usepackage{bm}
\usepackage{xcolor}
\usepackage{algorithm}
\usepackage{algpseudocode}
\usepackage{float}
\usepackage{multirow}
\usepackage{booktabs}
\usepackage{lscape}

\usepackage{graphicx}
\usepackage{subcaption}
\usepackage{mwe}
\usepackage{url}
\usepackage{rotating} 

\newcommand{\basic}{Basic-PR-ELM}
\newcommand{\opt}{Opt-PR-ELM}
\newcommand{\seq}{S-R-ELM}

\ifCLASSINFOpdf
\else
\fi

\begin{document}
%
\title{An Optimized and Energy-Efficient Parallel Implementation of Non-Iteratively Trained Recurrent Neural Networks}
%
%
%

\author{Julia El Zini,~\IEEEmembership{Member,~IEEE,}
         Yara Rizk,~\IEEEmembership{Member,~IEEE,}
        and Mariette Awad,~\IEEEmembership{Member,~IEEE}
        \IEEEcompsocitemizethanks{\IEEEcompsocthanksitem J. El Zini Y. Rizk and M. Awad are with the Department of Electrical and Computer Engineering, American University of Beirut, Beirut, Lebanon.\protect \\
        	E-mail: {jwe04,yar01,mariette.awad}@aub.edu.lb}
}

%
%


\markboth{Journal of \LaTeX\ Class Files,~Vol.~14, No.~8, August~2015}%
{Shell \MakeLowercase{\textit{et al.}}: Bare Demo of IEEEtran.cls for IEEE Journals}
%



\maketitle


\begin{abstract}
	Recurrent neural networks (RNN) have been successfully applied to various sequential decision-making tasks, natural language processing applications, and time-series predictions. Such networks are usually trained through back-propagation through time (BPTT) which is prohibitively expensive, especially when the length of the time dependencies and the number of hidden neurons increase. To reduce the training time, extreme learning machines (ELMs) have been recently applied to RNN training, reaching a 99\% speedup on some applications. Due to its non-iterative nature, ELM training, when parallelized, has the potential to reach higher speedups than BPTT. 
	
	In this work, we present \opt, an optimized parallel RNN training algorithm based on ELM that takes advantage of the GPU shared memory and of parallel QR factorization algorithms to efficiently reach optimal solutions.  The theoretical analysis of the proposed algorithm is presented on six RNN architectures, including LSTM and GRU, and its performance is empirically tested on ten time-series prediction applications. \opt~is shown to reach up to $845$  times speedup over its sequential counterpart and to require up to $20$x less time to train than parallel BPTT.

\end{abstract}


%
\IEEEpeerreviewmaketitle

\section{Introduction}
Recurrent neural networks (RNN) are a type of neural networks that have been successfully applied to many problems in machine learning \cite{lecun2015deep}. They have proven their ability to exceed human performance in time series prediction and sequential decision-making \cite{schmidhuber2015deep}. RNNs' training is usually based on gradient descent methods, specifically back-propagation through time (BPTT) \cite{werbos1990backpropagation},  and real-time recurrent learning \cite{williams1995gradient} which require a substantial amount of iterations before converging. Moreover, when unfolded through time, RNNs become even deeper \cite{bengio1994learning} and their training becomes even more expensive since the number of learned weights grows exponentially with the number of hidden neurons and the length of time dependency. 

Non-iterative training algorithms have been investigated in the literature \cite{schmidt1992feedforward,bengio1994learning,te1995random} to reduce the training cost of neural networks. Recently, Ertugrul et al. \cite{ertugrul2016forecasting} proposed a non-iterative training algorithm for Jordan RNNs\cite{jordan1997serial}. Then, Rizk et al. \cite{rizk2019extreme} extended it to different RNN architectures, including Elman, fully connected RNN, and Long Short-Term Memory (LSTM). Their algorithm was tested on time-series and sequential decision-making problems and achieved a speedup of up to 99\% over iterative training. 

Although they only need one iteration to obtain near-optimal solutions, non-iterative training algorithms minimize their cost function by computing a Moore-Penrose pseudo-inverse which requires ample computational resources, especially for large matrices. To the best of our knowledge, no attempts have been made in the literature to parallelize non-iterative training algorithms for RNNs. Fortunately, such algorithms hold great potential for parallelization due to their non-sequential nature. 

In this work, we propose \basic, a basic parallel version of ELM training applied on six RNN architectures: Elman, Jordan, NARMAX, fully connected, LSTM, and GRU. \basic~relies on parallel QR factorization to solve the pseudo-inverse required in ELM training algorithms. Then, the memory access patterns were studied and led to \opt, an optimal version of parallel ELM training that utilizes the GPU shared memory to speedup up the training process further. 

The proposed algorithms, \basic~and \opt, are tested on $10$ publicly available time-series prediction applications and on different GPU architectures to empirically show their scalability, robustness, portability, speedup potentials, and energy efficiency. Compared to the sequential version proposed by Rizk et al. in \cite{rizk2019extreme}, \basic~and \opt~achieve a speedup of up to $599$ and $845$, respectively while consuming less power. Notably, \opt~is shown to train LSTM networks $20$ times faster than the parallel iterative training algorithms (BPTT).

The rest of the paper is organized as follows: Section~\ref{sec:background} presents the background on ELM-training and the RNN architectures. Section~\ref{sec:related_work} summarizes the related work on RNN training and the parallel training algorithms. Section~\ref{sec:metho} presents the proposed algorithms \basic~and \opt~and Section~\ref{sec:theo} theoretically analyzes their memory and floating-point operations. Then, Sections~\ref{sec:exp_setup} discusses the experimental setup and Section~\ref{sec:exp_results} reports the empirical results. Finally, Section~\ref{sec:conc} concludes with final remarks. 

%
%
%
%
%
%
\section{Background}\label{sec:background}
\subsection{Extreme Learning Machine}
Extreme Learning Machine (ELM) is a non-iterative training algorithm introduced by Huang et al.  \cite{huang2004extreme} for single hidden layer feedforward neural networks (SLFNs).
Given $n$ arbitrary distinct training samples $(\bm{x}_j, y_j)$ where $\bm{x}_j \in \mathbb{R}^m, y_j \in \mathbb{R}$, $M$ hidden nodes and $g$ as activation function, the predicted output $O_j $ can be written as 
$\sum_{i = 1}^{M}\beta_i g(\bm{w}_i^\mathsf{T}\bm{x}_j + b_i) $
where $\bm{w}_i \in \mathbb{R}^m$ is the weight vector connecting the $i$th hidden node and the input nodes, $\bm{\beta}\in\mathbb{R}^M$ is the weight vector connecting all the hidden nodes and the output node and $b_i $ is the bias of the $i^{\text{th}}$ hidden node. Throughout the training, the input weights $w_{ij}$ are randomly generated and fixed and the output weights $\beta_1 \dots \beta_M$ are analytically computed. 
The goal is to minimize the error between the predicted and the true output as:
\begin{equation}\label{min1}
\min_{\beta} \sum_{j=1}^n \lVert O_j - t_j\rVert^2 = \sum_{j = 1}^{n} \lVert \sum_{i = 1}^{M}\beta_i g(\bm{w}_i^\mathsf{T}\bm{x}_j + b_i)  - t_j\rVert
\end{equation}
Defining $\bm{H}$ and $\bm{T}$ as:
\begin{equation}\label{eq:h_comp}
\bm{H}_{(n\times M)}
=
\begin{bmatrix}
g(\bm{w}_1^\mathbf{T}\bm{x}_1 + b_1) & \dots   & g(\bm{w}_M^\mathbf{T}\bm{x}_1 + b_M) \\
\vdots & \ddots & \vdots\\
g(\bm{w}_1^\mathbf{T}\bm{x}_n + b_1) & \dots   & g(\bm{w}_M^\mathbf{T}\bm{x}_n + b_M) \\
\end{bmatrix}
\end{equation}

\begin{equation}
\bm{T}_{(n\times 1)}
=[
t_1, t_2 , \dots, t_n ]^{\mathsf{T}},
\end{equation}
one can compactly write the problem in Eq.~\ref{min1} as minimizing $\lVert    \bm{H\beta}-\bm{T}\rVert^2$. The solution of this problem is given as: $\bm{\beta} = \bm{H}^\dagger \bm{T}$, where $\bm{H}^\dagger = (\bm{H}^{\mathsf{T}}\bm{H})^{-1}\bm{H}$ is the Moore-penrose generalized inverse of the matrix $\bm{H}$. 
\vspace{-0.9em}
\subsection{RNN architectures}\label{sec:bckg_arch}
\begin{figure}
	\includegraphics[width=0.49\textwidth]{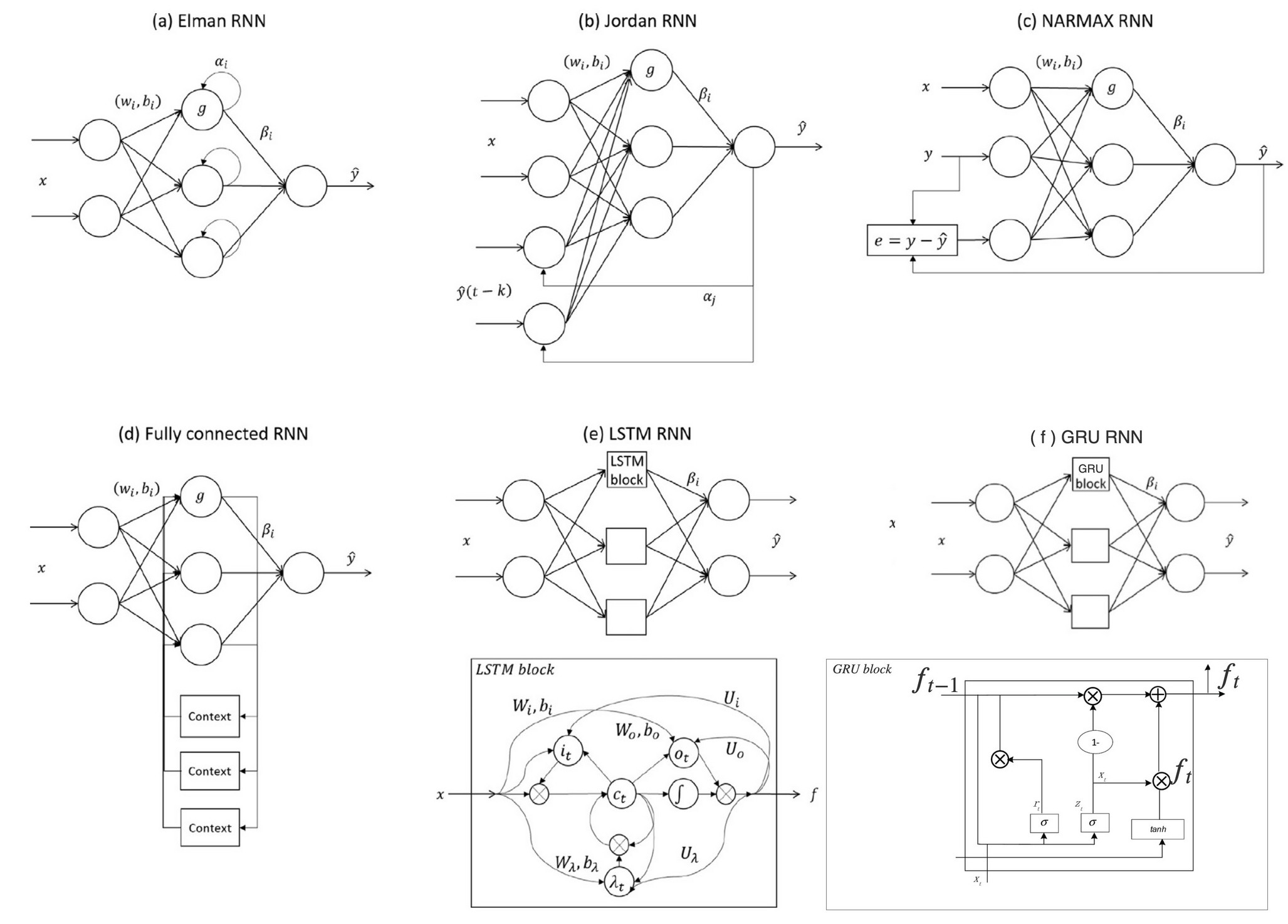}
	\caption{RNN architectures adapted from prior work in \cite{rizk2019extreme}}\label{fig:arch}
\end{figure}

RNNs are one of the most powerful neural networks that are best suitable to model long-term dependencies in time-series applications \cite{schmidhuber2015deep}. RNN architectures differ in the way cycles are introduced in the network. In this work, we consider six RNN architectures, illustrated in Fig.\ref{fig:arch}: Elman \cite{elman1990finding}, Jordan \cite{jordan1997serial}, NARMAX \cite{connor1994recurrent}, fully connected RNN, LSTM \cite{hochreiter1997long} and GRU \cite{cho2014learning}. 

In Fig.\ref{fig:arch} and throughout this work, $ \bm{x}\in S \times Q $ is the input to the network, $M$ is the number of hidden neurons, $\bm{w}_i \in \mathbb{R}^S$ is the vector of weights connecting the input to the $i^{th}$  neuron, $\alpha_{ik} \in \mathbb{R}$ is the weight from the neuron $i$ to itsef from the $k^{th}$ previous time step and $b_i$ is $i^{\text{th}}$ bias.
\vspace{-0.5em}
\subsubsection{Elman}
Elman RNNs are single hidden layer networks where context neurons introduce recurrence by feeding back signals as internal state of the network. At time step $t$, the output is: 
\begin{equation}\label{eq:yhat_elman_jordan}
\hat{y} = \sum_{i=1}^{M} \beta_i f_i(t)
\end{equation}
where $f_i(t) = g\bigg(\bm{w}_i^{\mathsf{T}} \bm{x}(t) + \sum_{k=1}^{Q} \alpha_{ik} f_i(t-k) + b_i\bigg)$ is the output of neuron $i$ at time $t$. 

\subsubsection{Jordan}
Jordan networks are similar to Elman's except for the way recurrence is introduced. In the Jordan architecture, signals are fed back from the predicted output of the previous time step. Consequently, such networks are more suitable for time series prediction where dependencies are on current input and previous outputs. Specifically, the output at time step $t$ is described by Eq.~\ref{eq:yhat_elman_jordan} with \\$f_i(t) = g\bigg(\bm{w}_i^{\mathsf{T}} \bm{x}(t) + \sum_{k=1}^{Q} \alpha_{ik} \hat{y}(t-k) + b_i\bigg)$. 

\subsubsection{NARMAX}
The \textbf{N}onlinear \textbf{A}utoreg\textbf{R}essive \textbf{M}oving \textbf{A}verage model with e\textbf{X}ogenous inputs (NARMAX) represents a wide class of nonlinear systems \cite{billings2013nonlinear}. NARMAX networks, have been proposed for non-linear time series prediction using artificial neural networks and are described by $\hat{y}(t) = \sum_{i=1}^{M} \beta_i g \bigg( \bm{w}_i^{\mathsf{T}} \bm{x}(t) + \sum_{l=1}^{F} w^{'}_{il}y(t-l) + \sum_{l=1}^{R}  w^{''}_{il}e(t-l) + b_i\bigg)$,
where $F$ and $R$ are the lengths of the time dependency of the output and the error feedbacks respectively, $e(t) = y(t) - \hat{y}(t)$, $w^{'}_{il} \in \mathbb{R}$ ($w^{''}_{il} \in \mathbb{R}$ resp.) is the weight from the output (error resp.) at the $l^{th}$ time step to the $i^{th}$ hidden neuron.

\subsubsection{Fully Connected RNN}
A fully connected RNN is the most general RNN architecture in which signals are fed back from all hidden neurons at previous time steps. Specifically, the output at time step $t$ is described by Eq.~\ref{eq:yhat_elman_jordan} with $f_i(t) = g\bigg(\bm{w}_i^{\mathsf{T}} \bm{x}(t) + \sum_{l=1}^{Q} \sum_{m=1}^{M}\bm{\alpha_}{ilk} f_m(t-k) + b_i\bigg)$. 
In this case, $\alpha_{ilk} \in \mathbb{R}$ is the weight connecting the neuron $i$ to neuron $l$ from the $k^{th}$ previous time step. 
\subsubsection{LSTM}
LSTMs were introduced by \cite{hochreiter1997long} to solve the vanishing gradient problem in BPTT. LSTMs have been successfully applied to a wide variety of applications inluding speech recognition \cite{graves2013hybrid,graves2013speech}, machine translation \cite{cho2014properties,wu2016google} and human action recognition \cite{liu2016spatio,liu2017skeleton}. An LSTM unit is composed of the main cell, an input, output and forget gates which regulate the flow of information into and out of the cell through forgetting factors and weights. This formulation gives the network the ability to decide on which information to remember. The output of LSTM is described by Eq.~\ref{eq:yhat_elman_jordan} with $f(t) = o(t) \circ g_f(c(t)) $, 
$\circ$ is the Hadamard product of two matrices and $o(t)$, $c(t)$, $\lambda(t)$ and $in(t)$ are given by:
\begin{equation*}\label{eq:o_lstm}
o(t) = g_o\big(W_ox(t) + U_o f(t-1) + b_o\big)
\end{equation*}
\begin{equation*}\label{eq:c_lstm}
c(t) = \lambda(t) \circ c(t-1) + in(t) \circ g_c\big(W_cx(t) + U_cf(t-1) + b_c\big)
\end{equation*}
\begin{equation*}\label{eq:lambda_lstm}
\lambda(t) = g_\lambda\big(W_\lambda x(t) + U_\lambda f(t-1) + b_\lambda\big)
\end{equation*}
\begin{equation*}\label{eq:in_lstm}
in(t) = g_{in}\big(W_{in} x(t) + U_{in} f(t-1) + b_{in}\big)
\end{equation*}

\subsubsection{GRU}
GRUs are introduced in \cite{cho2014learning} as a gating mechanism for RNNs. They resemble LSTMs but have only two gates and fewer parameters. GRUs expose their state at each time step and do not have any mechanism to control the degree to which their state is exposed \cite{chung2014empirical}. They exhibit good performances on small datasets \cite{chung2014empirical} and are widely used in speech recognition \cite{tang2017memory,chorowski2015attention} and sequence modeling \cite{chung2014empirical}. 
GRUs' output is described by Eq.~\ref{eq:yhat_elman_jordan} while $f(t)$ is given by: 
\begin{multline}
f(t) = \big(1-z(t)\big) \circ f(t-1) + z(t) \circ g_f\big(W_fx(t) +\\U_f(r_t \circ f(t-1 ) +b_f)\big)
\end{multline}
where $z(t) = g_z(W_zx(t) + U_zf(t-1) + b_z)$ and $ r(t) = g_r(W_rx(t) + U_rf(t-1) + b_r)$.

\section{ Related Work }\label{sec:related_work}
This work focuses on the parallelization of a non-iterative training algorithm for RNNs. In what follows, we first discuss the basic training methods of RNNs while focusing on the non-iterative ones. Then, we report the parallelization attempts for training algorithms.

\subsection{RNN Training}
\subsubsection{Iterative RNN Training}
Training RNNs has been mainly done iteratively through BPTT \cite{werbos1990backpropagation} which unfolds the recurrence through time to transform the RNN into a feedforward network trained using gradient descent. BPTT is susceptible to local minima and suffers from the vanishing and exploding gradient problems with long time dependencies. BPTT can also be slow, given that it is applied iteratively in batch mode. Other iterative algorithms include, but are not limited to, Hessian free optimization \cite{martens2011learning}, extended Kalman filters \cite{wang2011convergence} and genetic algorithms (GA) \cite{blanco2001real}. Although successful, these algorithms are computationally expensive and require manually tuning of many hyper-parameters.  

\subsubsection{Non-Iterative RNN Training }
Different non-iterative training algorithms have been proposed to reduce the computational cost of training neural networks in general. For instance, the authors in \cite{schmidt1992feedforward,te1995random,pao1994learning,huang2004extreme} proposed ELM, a non-iterative method to train single hidden layer feedforward networks by randomly assigning input weights and computing output weights using the least-squares method. These methods were later extended to RNN architectures when Ertugrul implemented a non-iterative training for the Jordan RNN architecture in electricity load forecasting applications \cite{ertugrul2016forecasting}. Later, Park et al. extended it to online RNNs \cite{park2017online} and Rizk et al. generalized the approach to more powerful RNN architectures \cite{rizk2019extreme}. 

Although these methods achieved high speedups (up to 99\% in \cite{rizk2019extreme}), they heavily rely on stencil operations and on the computation of the generalized inverse of matrices which are CPU intensive operations and could be further optimized using parallel algorithms.  

\subsection{Parallelizing  Training Algorithms}
Several frameworks have been developed to solve challenges of high performance computing in the big data area \cite{zhang2019guest}, including parallelizing training algorithms. This is the first attempt to parallelize non-iterative training of RNNs; thus we describe previous work on the parallelization of RNN \textit{iterative} training algorithms and on the \textit{parallel} \textit{non-iterative} training for neural networks - not exclusively RNN. 

\subsubsection{Parallelizing Iterative Training Algorithms For RNN}
Parallelizing RNN training is mostly based on parallelizing the back-propagation algorithm (BP). For instance, Sierra et al. parallelized BP on CUBLAS and achieved a speedup of $63$. In \cite{zhang2015parallel}, data is distributed on multiple GPUs achieving a speedup of up to $51$ \cite{sierra2010parallel}. In \cite{wang2019scaling}, parallel scan algorithm improves the step complexity of BP from $\mathcal{O}(n)$ to $\mathcal{O}(\log n)$. Khomenko et al. parallelized their data on multiple GPUs and relied on batch bucketing by input sequence length to accelerate RNN training achieving a speedup of up to $4$  \cite{khomenko2016accelerating}. In \cite{ouyang2017fast}, a semantic correlation-based data pre-fetch framework is implemented to break the dependency in the input to parallelize the training of cognitive applications \cite{ouyang2017fast}. Their work is tested on LSTMs using image captioning, speech recognition, and language processing applications showing a speedup of $5.1$, $44.9$ and $1.53$, respectively. Recently, GA is introduced into the Elman architecture to accelerate the training and prevent the local minima problem \cite{jia2019novel}. GA-Elman outperformes traditional training algorithms in terms of convergence speed and accuracy.

\subsubsection{Parallelizing Non-Iterative Training Algorithms}
Non-iterative training algorithms for RNNs are shown to require less training time than iterative methods \cite{rizk2019extreme,ertugrul2016forecasting,park2017online}. However, even with non-iterative training, large datasets require costly computations, especially when increasing the number of neurons or when model selection is performed to avoid over-fitting \cite{van2011gpu}. Parallelizing non-iterative training has been explored in single layer feedforward networks by \cite{he2013parallel}. Their approach is based on a Map-Reduce and achieves a speedup of up to $5.6$ when tested on $32$ cores. Following a similar approach, Wang et al. \cite{wang2015parallel} developed a parallel implementation of online ELM and achieved a speedup of $3.5$ when trained on $120$K instances with $120$ attributes. Huang et al. extended their approach to the ensemble online sequential ELM which was tested on real and synthetic data with $5120$K training instances and $512$ attributes and achieved a speedup of $40$ on a cluster with $80$ cores \cite{huang2016parallel}. In \cite{van2011gpu}, Van et al. attempted to parallelize ELM on Flink with multi hidden layer feedforward network and achieved a speedup of $17$ . 

To the best of our knowledge, our work is the first attempt to parallelize non-iterative training for different RNN architectures.

\section{Methodology}\label{sec:metho}
Before proposing our methods, we present the nomenclature that will be used throughout this paper in Table~\ref{nomen}. 
\renewcommand{\arraystretch}{1.4}
\begin{table}[]
	\caption{Nomenclature} \label{nomen}
	\begin{tabular}{|p{2.9cm}p{5.2cm}|}
		\hline
		\textbf{Symbol} & \textbf{Definition} \\
		$n$ & Number of training samples \\
		$M$ & Number of hidden neurons \\
		$Q$ & Max number of time dependencies \\
		$S$ & Dimension of input \\
		$\bm{x}_j \in \mathbb{R}^{S \times Q}$ & $j^{th}$ Input instance  \\
		$y_j \in \mathbb{R}$ & $j^{th}$ Output instance \\ 
		$\bm{X} \in \mathbb{R}^{n \times S \times Q}$ & Input matrix \\        
		$\bm{Y} \in \mathbb{R}^{n}$ & Output matrix \\
		$\bm{W} \in \mathbb{R}^{S \times L}$ & Weight matrix connecting the input to the hidden neurons\\
		$\bm{\alpha} \in \mathbb{R}^{L \times Q} $ & Weight matrix connection the hidden neuron to itself for previous time steps \\
		$\bm{b} \in \mathbb{R}^{L}$ & Bias vector for the hidden neurons \\         
		$\bm{\beta} \in \mathbb{R}^{L}$ & Weight vector connecting hidden neurons to output layer \\ 
		\seq & Sequential ELM for RNN training  \\
		\basic & Basic parallel ELM RNN training \\
		\opt & Optimal parallel ELM RNN training \\
		BPTT & Back-propagation through time \\
		P-BPTT & Parallel Back-propagation through time \\
		BS & Block size \\
		TW & Tile width\\
		\hline
	\end{tabular}
\end{table}
In this work, a parallel version of ELM-trained RNNs will be formalized and implemented. The sequential version of our approach, denoted by S-RELM, is summarized in algorithm~\ref{algo:rnn_elm} and is adopted from our previous work in \cite{rizk2019extreme}.
\begin{algorithm}[!ht]
	\caption{\seq~algorithm}\label{algo:rnn_elm}
	\begin{algorithmic}[1]
		\State Randomly assign $\bm{W}, \bm{\alpha}, \bm{b}$
		\State Compute $\bm{H}(t), t=1\dots Q$ according to the corresponding RNN architecture
		\State Compute $\bm{\beta} = \bm{H}(Q)^\dagger \bm{Y}$ using the generalized Moore–Penrose pseudoinverse
	\end{algorithmic}
\end{algorithm}

$\bm{H}$(t) at row $i$ and column $j$ is referred to as $h_{ij}[t]$ in this paper and is computed as in Equations~\ref{eq:hij_elman}, \ref{eq:hij_jordan}, \ref{eq:hij_narmax}, \ref{eq:hij_fully}, \ref{eq:hij_lstm} and \ref{eq:hij_gru} for the Elman, Jordan, NARMAX, fully connected, LSTM and GRU architectures respectively. 
\begin{equation}\label{eq:hij_elman}
h_{ij}[t]= g(\bm{W}[:,j] . \bm{X}[i,:,t] + \bm{b}_i + \sum_{k=1}^{Q} \bm{\alpha}[j,k] h_{ij}[t-k]
\end{equation}
\begin{equation}\label{eq:hij_jordan}
h_{ij}[t]= g(\bm{W}[:,j] . \bm{X}[i,:,t] + \bm{b}_i + \sum_{k=1}^{Q} \bm{\alpha}[j,k] \hat{y}(t-k)
\end{equation}
\begin{multline}\label{eq:hij_narmax}
h_{ij}[t] = g(\bm{W}[:,j] . \bm{X}[i,:,t] + \bm{b}_i + \sum_{l=1}^{F} \bm{W'}[i,l]  y(t-l) +\\ \sum_{l=1}^{R} \bm{W''}[i,l]  e(t-l)
\end{multline}
\begin{multline}\label{eq:hij_fully}
h_{ij}[t] = g\bigg(\bm{W}[:,j] . \bm{X}[i,:,t] + \bm{b}_i +\\ \sum_{k=1}^{Q} \sum_{l=1}^{M}\bm{\alpha}[j,l,k]  h_{ij}[t-k]\bigg)
\end{multline}
\begin{multline}\label{eq:hij_lstm}
h_{ij}[t] = \bm{o}[i,j,t] \circ g_f\big(\bm{c}[i,j,t]\big)\\
\end{multline}
\begin{multline}\label{eq:hij_gru}
h_{ij}[t]= \bigg(1-\bm{z}[i,j,t]\bigg) \circ h_{ij}[t-1] + \bm{z}[i,j,t] \circ \\g_f\bigg(\bm{W_f}[:,j] . \bm{X}[i,:,t] +U_f(\bm{r}[i,j,t] \circ h_{ij}[t-1 ] + \bm{b}_i )\bigg)
\end{multline}

Considering Algorithm~\ref{algo:rnn_elm}, one can see that the running time of the ELM training mainly consists of two CPU intensive operations: computing $\bm{H}$ and computing $\bm{\beta}$ by solving the linear system using the Moore-Penrose pseudo-inverse. Thus, those two operations are the main target when optimizing the performance of non-iterative training. 

\subsection{$\bm{H}$ Computation}
\subsubsection{Basic Parallel Implementation (Basic-PR-ELM)}
For all RNN architectures, the computation of $\bm{H}(t)$ at row $i$ and column $j$ is independant of the computation of $\bm{H}(t)$ at row $i_2$ and column $j_2$, $\forall i_2 \neq i, j_2\neq j$; it only depends on $\bm{H}(t_2)$ at row $i$ and column $j$ for $t_2 < t$. Given only this dependency, a parallel $\bm{H}$ computation can be done as follows: each thread $(i,j)$ can independently compute $\bm{H}(t)$ at row $i$ and column $j$ for $t=1,\dots,Q$. We describe the basic implementation of the computation of $\bm{H}$  for the Elman architecture in Algorithm~\ref{algo:H_basic}. 
\begin{algorithm}[!ht]
	\caption{Basic-PR-ELM by thread $(i,j)$}\label{algo:H_basic}
	\begin{algorithmic}[1]
		
		\State $tx \gets threadIdx.x$
		\State$ty \gets cuda.threadIdx.y$
		
		\State$Row \gets tx + blockIdx.x \times blockDim.x$ 
		\State$Col \gets ty + blockIdx.y \times blockDim.y$ 
		\For{$t=1 \mapsto Q$ }
		
		\State $h_{ij}\gets \bm{W}[:,Col] . \bm{X}[Row,:,t]$ 
		\State $h_{ij} \gets h_{ij} + b_{Col}$
		\For{$t_{prev}=1$ $\mapsto$ $t$ }
		
		\State  $h_{ij} \gets h_{ij} + \alpha[j,t_{prev}] \times H[Row, Col, t_{prev}]$  
		\EndFor
		\State $H[Row, Col, t]\gets h_{ij}$
		\EndFor
	\end{algorithmic}
	
\end{algorithm}
\subsubsection{Optimized Parallel Implementation (Opt-PR-ELM)}
\begin{figure}
	\centering
	\includegraphics[width=0.45\textwidth]{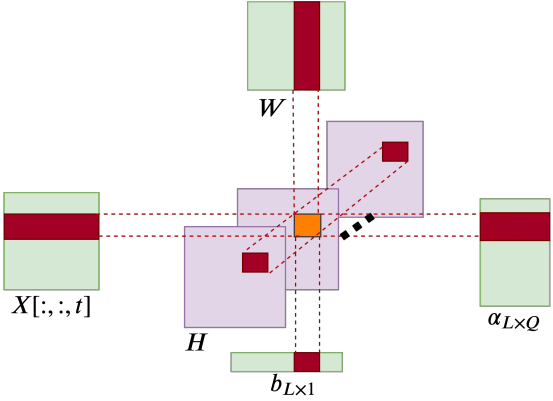}
	\caption{\basic~memory access patterns on Elman}\label{fig:mem_access}
\end{figure}

Fig.~\ref{fig:mem_access} illustrates the memory access patterns of \basic~on the Elman architecture. One can clearly see that threads in the same row access the same elements of $\bm{X}$ and threads in the same column access the same elements of $\bm{W}$ and $\bm{\alpha}$. Thus, the tiling concept can be applied to utilize the shared memory to speed up the computation of $\bm{H}$. Moreover, we notice that $b_{Col}$ can be preloaded and used efficiently by other threads. Algorithm~\ref{algo:H_opt} describes how these optimizations can be applied for the Elman architecture. First, in the dot product $ \bm{W}[:,Col] . \bm{X}[Row,:,t]$, each thread can load only one element of $\bm{W}$ and one element of $\bm{X}$  into the shared memory. Once the threads synchronize, then all needed elements of $\bm{W}$  and $\bm{X}$ are loaded, and the dot product can be computed efficiently. Second, only one thread can load $\bm{b}[j]$ that is needed by all the threads in the same column of the block. The same tiling concept used to compute $ \bm{W}[:,Col] . \bm{X}[Row,:,t]$ can be used to speed up the computation of $\alpha[j,t_{prev}] \times H(t_{prev})[Row, Col]$. Lastly, each thread can save the values of $H(t)[Row, Col]$ in its register file to reduce the time taken to read from the global memory in line $8$ of Algorithm~\ref{algo:H_basic}. If these values do not fit in the registers, they are read from the global memory.  

Alogirhtms~\ref{algo:H_basic}~and~\ref{algo:H_opt} could be easily extended to other architectures when Eq.~\ref{eq:hij_elman} is replaced by Eq~\ref{eq:hij_jordan}, \ref{eq:hij_narmax}, \ref{eq:hij_fully}, \ref{eq:hij_lstm} or \ref{eq:hij_gru}.

\begin{algorithm}[!ht]
	\caption{\opt~by thread $(i,j)$}\label{algo:H_opt}
	\begin{algorithmic}[1]
		\State $tx \gets threadIdx.x$
		\State $ty \gets cuda.threadIdx.y$
		
		\State $Row \gets tx + blockIdx.x \times blockDim.x$ 
		\State $Col \gets ty + blockIdx.y \times blockDim.y$ 
		
		\State $\bm{H}_{loc} \gets t$-dimensional array in the local memory of thread $(i,j)$\;
		\For{$t\gets1 \mapsto Q$ }
		\State  $h_{ij}\gets 0$ 
		\For{$tile= 1 \mapsto num\_tiles$ }
		\State $\bm{W}_{\text{shared}} \gets \bm{W}[tx + tile \times TW:,Col]$ 
		\State $\bm{X}_{\text{shared}} \gets \bm{X}[Row, ty + tile * TW,t]$ 
		
		\State synch()
		
		\State $h_{ij} \gets h_{ij} + \bm{W}_{shared} . \bm{X}_{shared}$ 
		\EndFor
		\State synch()
		\If{$tx = 0$ and $ty=0$}
		\State $\quad b_{\text{shared}} \gets \bm{b}[Col]$
		\EndIf
		\State synch()
		\State $h_{ij} \gets h_{ij} + b_{\text{shared}}$
		\For{$tile\gets1 \mapsto \lceil \frac{t}{TW}\rceil$ }
		\State $\bm{\alpha}_{\text{shared}} \gets \bm{\alpha}[Col,tx + tile \times TW]$ 
		\State synch()
		\State $h_{ij} \gets h_{ij} + \bm{\alpha}_{\text{shared}} . \bm{H}_{loc}[t_{prev}]$  
		\EndFor
		\State synch()
		\State $ \bm{H}_{loc}[t] \gets h_{ij}$
		\State $ \bm{H}[Row, Col, t] \gets h_{ij}$
		\EndFor
		
	\end{algorithmic}
\end{algorithm}
\subsection{Computing $\bm{\beta}$ }
$\bm{\beta}$ is the solution of the following system: $\bm{H}\bm{\beta} = \bm{Y}$. Instead of comuting the pseudo-inverse $\bm{H}^{\dagger}$ and then multiplying it by $\bm{Y}$, one can perform a QR factorization of $\bm{H}$ as  $\bm{H} = \bm{Q}\bm{R}$, then compute $z= \bm{Q}^{\mathsf{T}}\bm{Y}$. Having that, $\bm{\beta}$ would be the solution of $\bm{R}\bm{\beta}=z$ by back substitution since $\bm{R}$ will be an upper triangular matrix. In this work, we make use of \textit{Numba} \cite{lam2015numba} and \textit{Numpy} \cite{numpy} libraries which provide an efficient implementation of this method in \textit{Python}.

\section{Theoretical Analysis}\label{sec:theo}
We analyze the memory read and write operations and the floating point operations (FLOPS) for the proposed algorithms: \basic~and \opt. For the Elman architecture, \basic~performs $Q(2S+Q+2)$ read operations divided as follows:
		\begin{itemize}
			\item$2\times SQ$ to read the values needed in line 6
			\item $Q$ reads for $b_{Col}$ in line 7
			\item $2\times\big(Q\frac{Q+1}{2}\big)$ reads in the loop at line 8
		\end{itemize}
	Moreover, only $Q$ write operations are needed (in line 11) and $Q(2S+Q+2)$ FLOPS are performed as follows:
	\begin{itemize}
		\item$2\times SQ$ to perform the dot product at line 6
		\item $Q$ FLOPS for the addition in line 7
		\item $2\times\big(Q\frac{Q+1}{2}\big)$ to perform the loop at line 8
	\end{itemize}

The memory operations to FLOPS ratio is $\frac{2S+Q+3}{2S+Q+2} > 1$ which might limit the performance of \basic. This ratio improves with \opt~ as it minimizes the memory operations while keeping the same number of FLOPS. Specifically, \opt~ decreases the number of reads to $\frac{1}{TW^2}\big(2\times SQ + \frac{Q(Q+1)}{2}\big)+1$ divided as follows:
	\begin{itemize}
		\item$\frac{2}{TW^2}\times SQ$ to read the values needed in line 12
		\item at most $1$ read for $b_{Col}$ in line 16
		\item $  \frac{1}{TW^2}\big(Q\frac{Q+1}{2}\big)$ reads in the loop at line 20
	\end{itemize}
where $TW$ is the tile width which is set to block size in this work. 
The new memory operations to FLOPS ratio is $\frac{\frac{1}{TW^2}\big(2\times SQ + \frac{Q(Q+1)}{2}\big)+1 + Q}{Q(2S+Q+2)}$ which is less then the ratio of \basic~ by a factor of $\approx TW^2$. Specifically, \opt~minimizes the number of read operations by a factor of $256$ ($1024$ resp.) when the tile width is set to $16$ ($32$ resp.).

Table~\ref{tbl:theo} reports the number of memory operations and FLOPS needed by \basic~ for each RNN architecture. The values of \opt~ are ommited as it requires the same number of write operations and FLOPS and less read operations by a factor of $\approx TW^2$.
\begin{table*}
	\centering
		\caption{Number of memory operations and FLOPS for each RNN architecture for \basic}\label{tbl:theo}
	\begin{tabular}{| l | c| c | c |}
		\hline
		\textbf{Architecture} & \textbf{\# Read Operations} & \textbf{\# Write Operations }& \textbf{FLOPS} \\
		\hline
		\textbf{Elman} & $Q\big(2S+Q+2\big)$ & $Q$ & $Q\big(2S+Q+2\big)$ \\
		\textbf{Jordan} & $Q\big(2S+1+(Q+1)(1/2+M)\big)$& $Q$& $Q\big(2S+1+\frac{Q+1}{2}(2SM+M)\big)$ \\
		\textbf{NARMAX} &$Q(2S +1 ) +2(2F+M+R)$ & $Q$ & $Q\big(2S+1 +2F+R(2+2SM+M)\big)$ \\
		\textbf{Fully Connected}& $Q\big(2S+1+2MQ\big)$& $Q$& $Q\big(2S+Q+2QM\big)$ \\
		\textbf{LSTM} & $Q(5S+13)$& $5Q$& $Q(8S+18)$\\
		\textbf{GRU} & $Q(4S+8)$ & $3Q$& $Q(3S+17)$\\
				\hline						
	\end{tabular}

\end{table*}
\section{Experimental Setup}\label{sec:exp_setup}
\subsection{Setup}
Serial algorithms were run on an Intel 64bit core-i5 machine with a memory of $8$ GB and $2133$ MHz. Parallel algorithms were run on NVidia Tesla K20m GPU with $2688$ CUDA cores and $723$MHz GPU core clock speed. The GPU main memory is $6$GB and bandwidth of $250$ GB/s between the host and the device. All experiments are repeated $5$ times, and the average value is reported. 

\subsection{Time Series Prediction Benchmarks}
Basic-PR-ELM and Opt-PR-ELM were validated on time series prediction problems. Table~\ref{tbl:db} presents the characteristics of the datasets ordered according to the number of instances. According to their size, we split the databases into three categories: small datasets containing less than $10$K instances, medium datasets with multiples of $10$K instances and large dataset consisting of multiples of $100$K instances. Japan population\footnote{kaggle.com/jd1325/japan-population-data}  tracks the population of various Japanese regions, while the Quebec Births\footnote{\url{datamarket.com/data/list/ ?q=provider\%3Atsdl}} tracks the number of births in Quebec and Exoplanet\footnote{\url{kaggle.com/keplersmachines/kepler-labelled-time-series-data}} describes the change in the light intensity of several thousand stars. Additionally, SP 500\footnote{\url{kaggle.com/benjibb/sp500- since-1950}} records the stock prices since 1950 while AEMO\footnote{\url{aemo.com.au/}} reports the electricity load demand in Australia  and hourly weather\footnote{\url{kaggle.com/selfishgene/historical-hourly-weather-data}} contains $\approx$ 5 years of temperature measures. The energy consumption dataset\footnote{\url{kaggle.com/selfishgene/historical-hourly-weather-data}} reports the hourly power consumption data in megawatts, the electricity load dataset\footnote{\url{archive.ics.uci.edu/ ml/index.php}} reports the electricity demand at the MT166 and MT257 substations and the stock prices dataset\footnote{\url{kaggle.com/camnugent/sandp500}} consists of historical stock prices for all companies currently on the S\&P 500 index. Finally, the temperature dataset\footnote{\url{kaggle.com/wkirgsn/electric-motor-temperature}} reports sensor data collected from a permanent magnet synchronous motor (PMSM) deployed on a testbench where PMSM represents a german OEM's prototype model.

\begin{table*}[]
	\caption{Benchmarks Description}\label{tbl:db}
	\begin{tabular}{ll ccc rrrr}
		\multicolumn{2}{c}{\textbf{Database}} & \multicolumn{3}{c}{\textbf{Size}} & \multicolumn{4}{c}{\textbf{Output Statistics}} \\
		\cmidrule(lr){1-2}\cmidrule(lr){3-5}\cmidrule(lr){6-9} 
		\textbf{Category}& \textbf{Name} & \textbf{\# of instances} & \textbf{Q} &\textbf{\% Train} & \textbf{Mean} &  \textbf{Std Dev} &  \textbf{Min} &  \textbf{Max} \\
		\hline
		\hline
		\textbf{Small} & Japan population & 2,540 & 10 & 80 & 1.40E+06 & 1.40E+06 & 1.00E+05 & 1.03E+08 \\
		& Quebec Births & 5,113 & 10 & 80 & 2.51E+02 & 4.19E+01 & -2.31E+01 & 3.66E+02 \\
		& Exoplanet & 5,657 & 3197 & 80 & -3.01E+02 & 1.45E+04 & -6.43E+05 & 2.11E+05 \\
		\hline
		\textbf{Medium} & SP500 & 17,218 & 10 & 80 & 8.99E+08 & 1.53E+09 & 1.00E+06 & 1.15E+10 \\
		& AEMO & 17,520 & 10 & 80 & 7.98E+03 & 1.19E+03 & 5.11E+03 & 1.38E+04 \\
		& Hourly weather & 45,300 & 50 & 80 & 2.79E+02 & 3.78E+01 & 0.00E+00 & 3.07E+02 \\
		\hline
		\textbf{Large} & Energy Consumption & 119,000 & 10 & 70 & 1.66E+03 & 3.02E+02 & 0.00E+00 & 3.05E+03 \\
		& Electricity load & 280,514 & 10 & 70 & 2.70E+14 & 2.60E+14 & 0.00E+00 & 9.90E+14 \\
		& Stock prices & 619,000 & 50 & 70 & 4.48E+06 & 1.08E+07 & 0.00E+00 & 2.06E+09 \\
		& Temperature & 998,000 & 50 & 70 & 5.07E+01 & 2.21E+01 & 4.00E+00 & 8.10E+01\\
		\hline
	\end{tabular}
	
\end{table*}

\section{Experimental Results}\label{sec:exp_results}

\subsection{Speedup}
\begin{figure*}
	\centering
	\begin{subfigure}[b]{0.95\textwidth}
		\centering
		\includegraphics[width=0.8\textwidth]{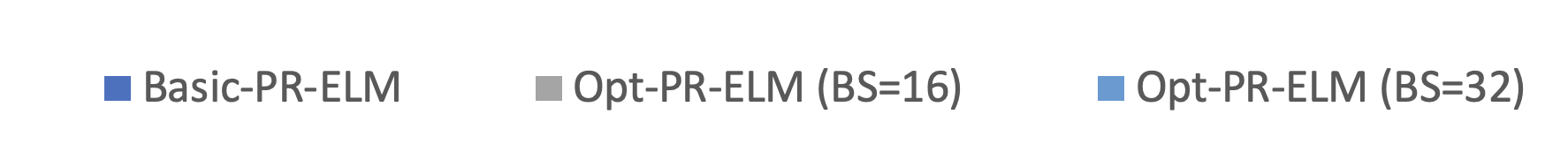}
	\end{subfigure}
	\begin{subfigure}[b]{0.475\textwidth}
		\centering
		\includegraphics[width=\textwidth]{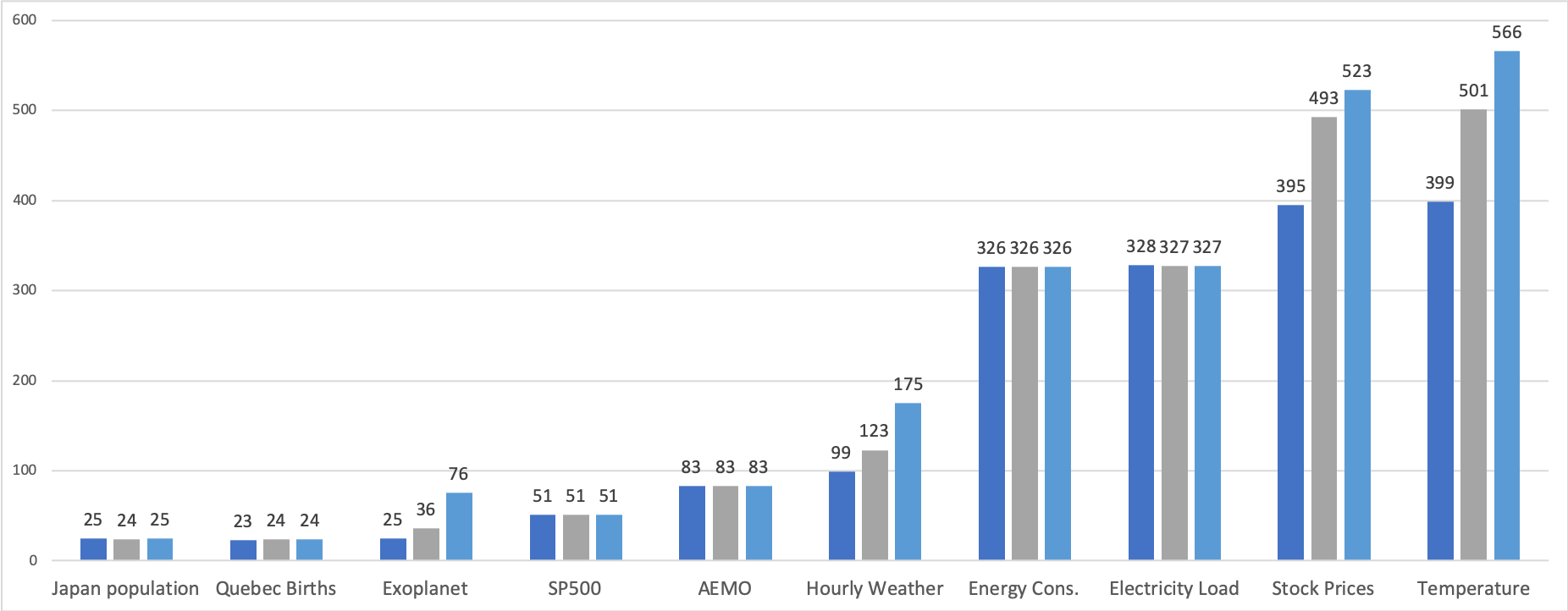}
		\caption{Jordan}
	\end{subfigure}
	\hfill
	\begin{subfigure}[b]{0.475\textwidth}  
		\centering 
		\includegraphics[width=\textwidth]{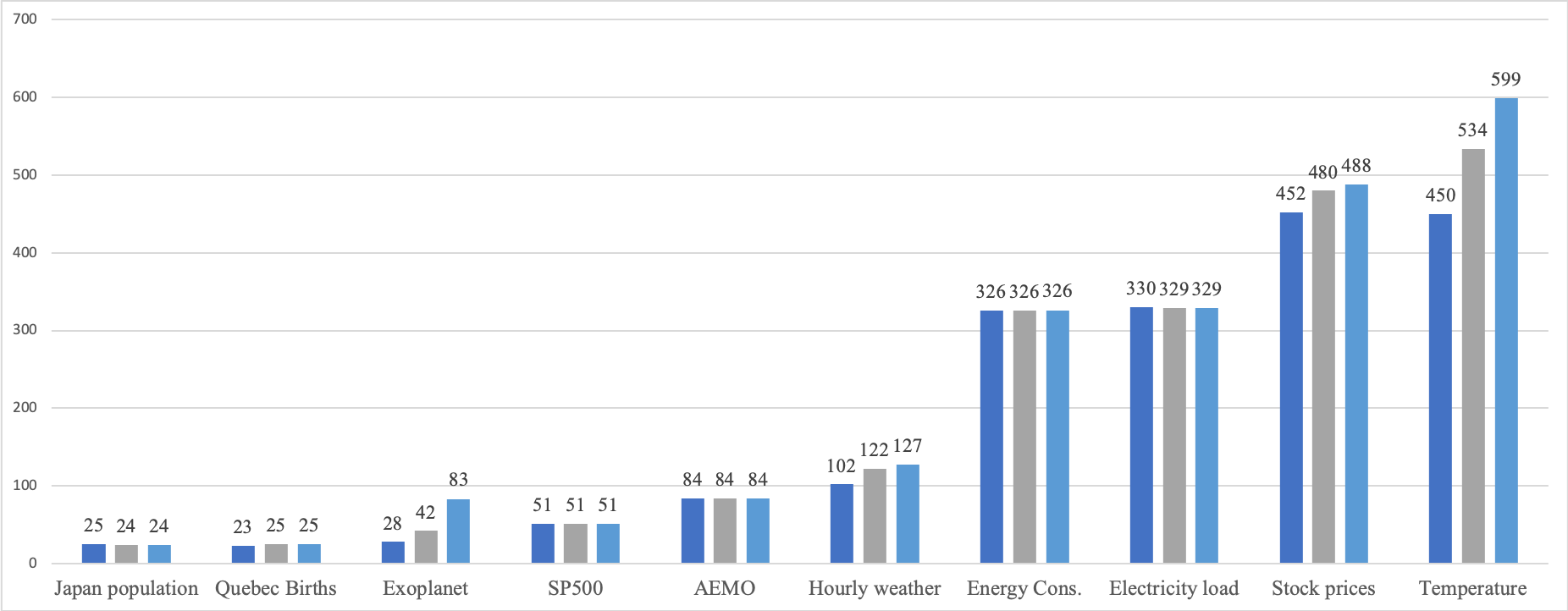}
		\caption{Elman}
	\end{subfigure}
	\vskip\baselineskip
	\begin{subfigure}[b]{0.475\textwidth}   
		\centering 
		\includegraphics[width=\textwidth]{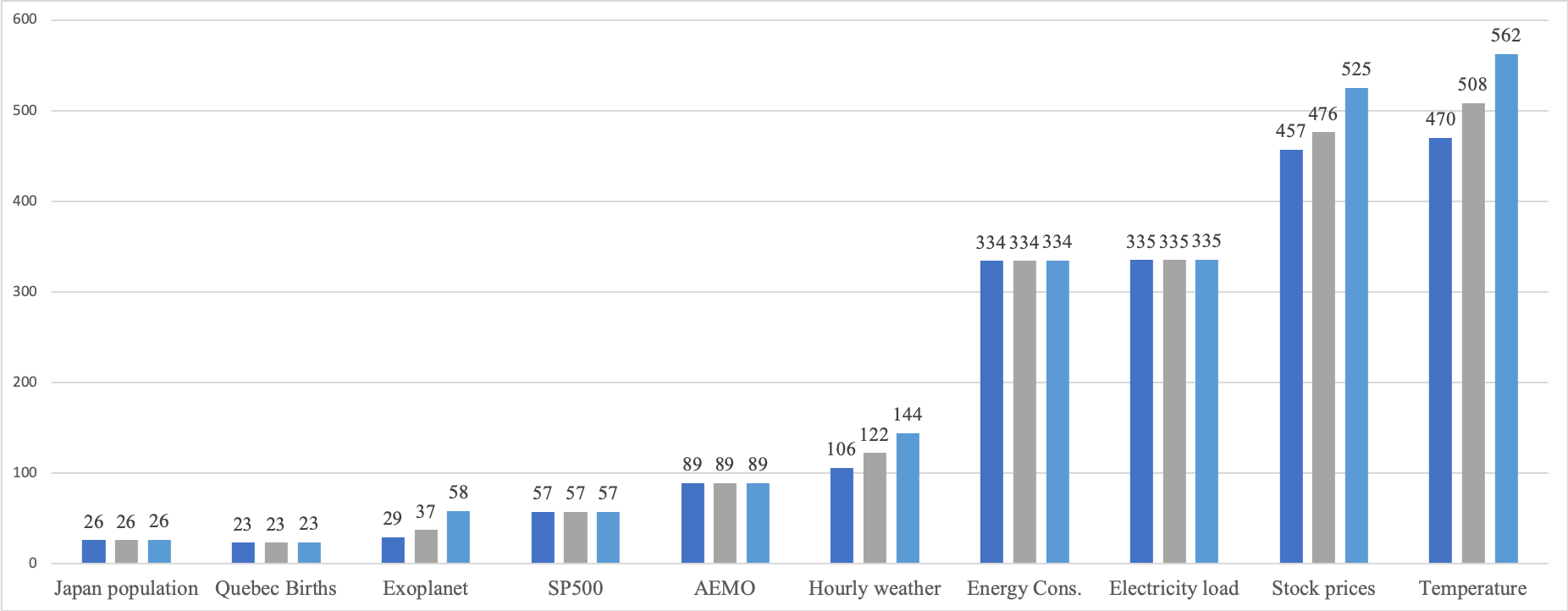}
		\caption{NARMAX}
	\end{subfigure}
	\quad
	\begin{subfigure}[b]{0.475\textwidth}   
		\centering 
		\includegraphics[width=\textwidth]{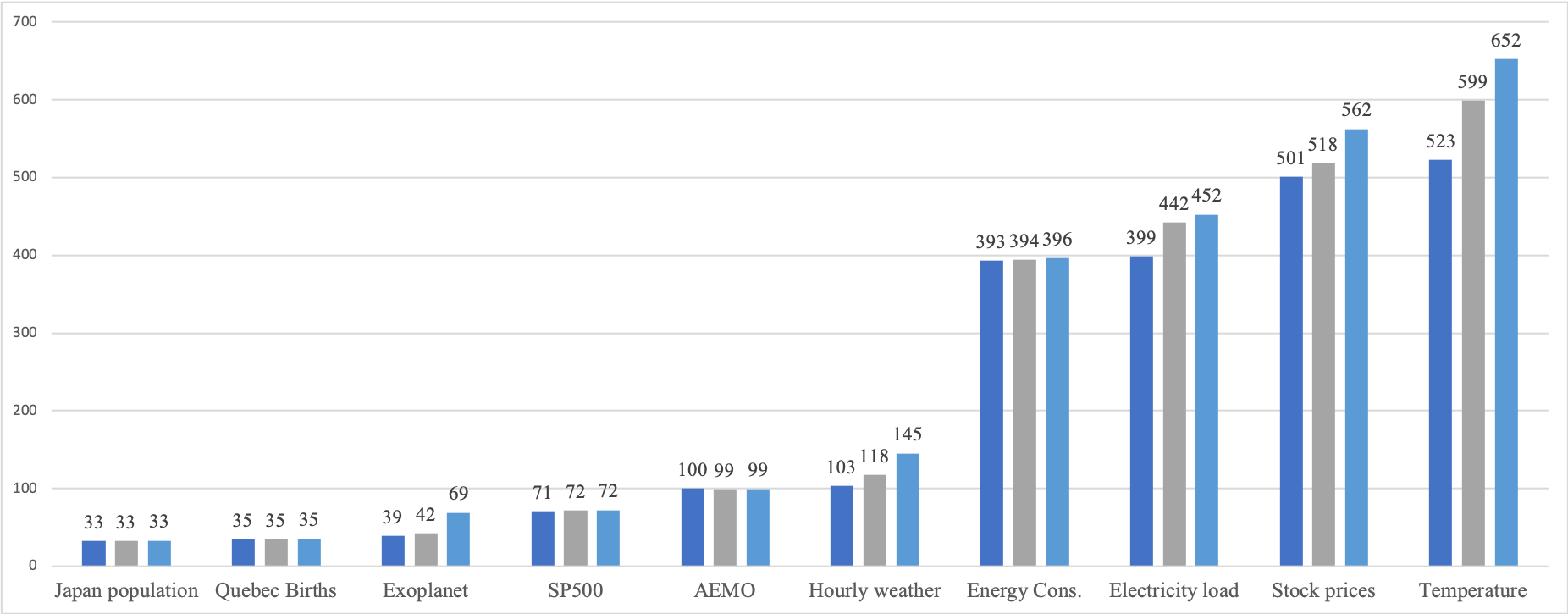}
		\caption{Fully Connected}
	\end{subfigure}
	\begin{subfigure}[b]{0.475\textwidth}   
		\centering 
		\includegraphics[width=\textwidth]{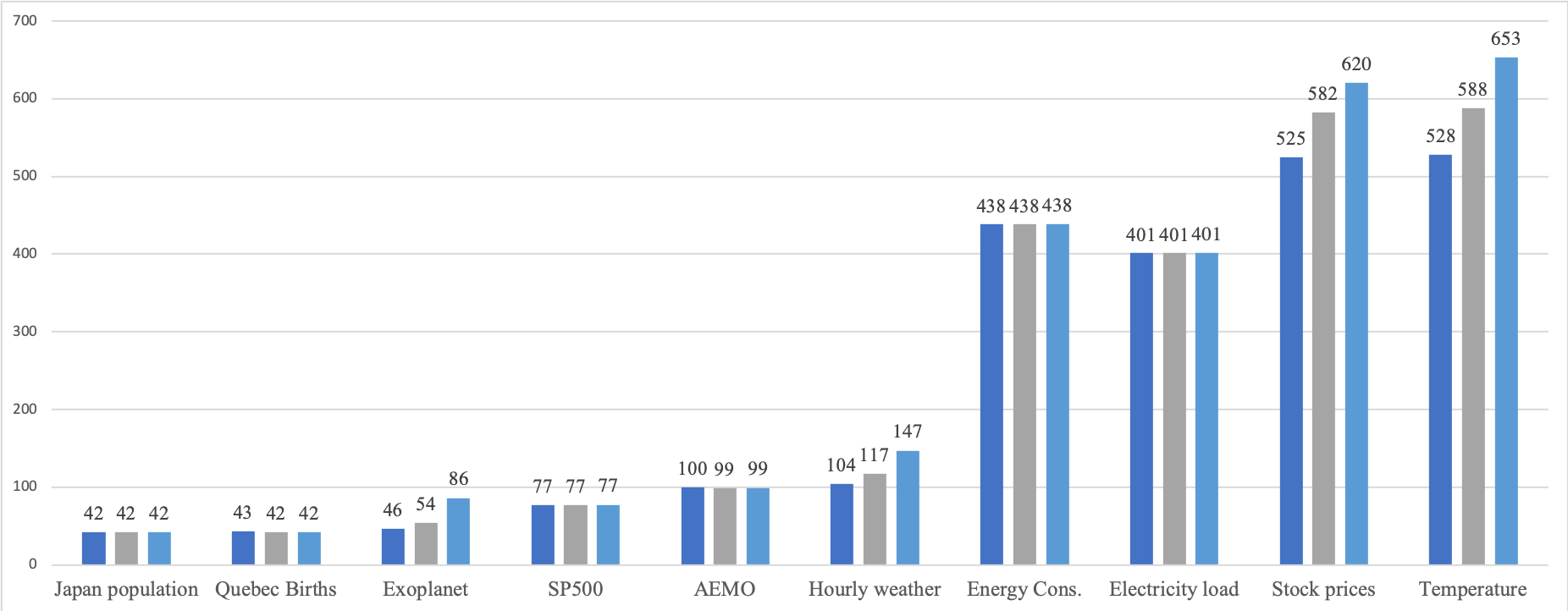}
		\caption{LSTM}
	\end{subfigure}
	\quad
	\begin{subfigure}[b]{0.475\textwidth}   
		\centering 
		\includegraphics[width=\textwidth]{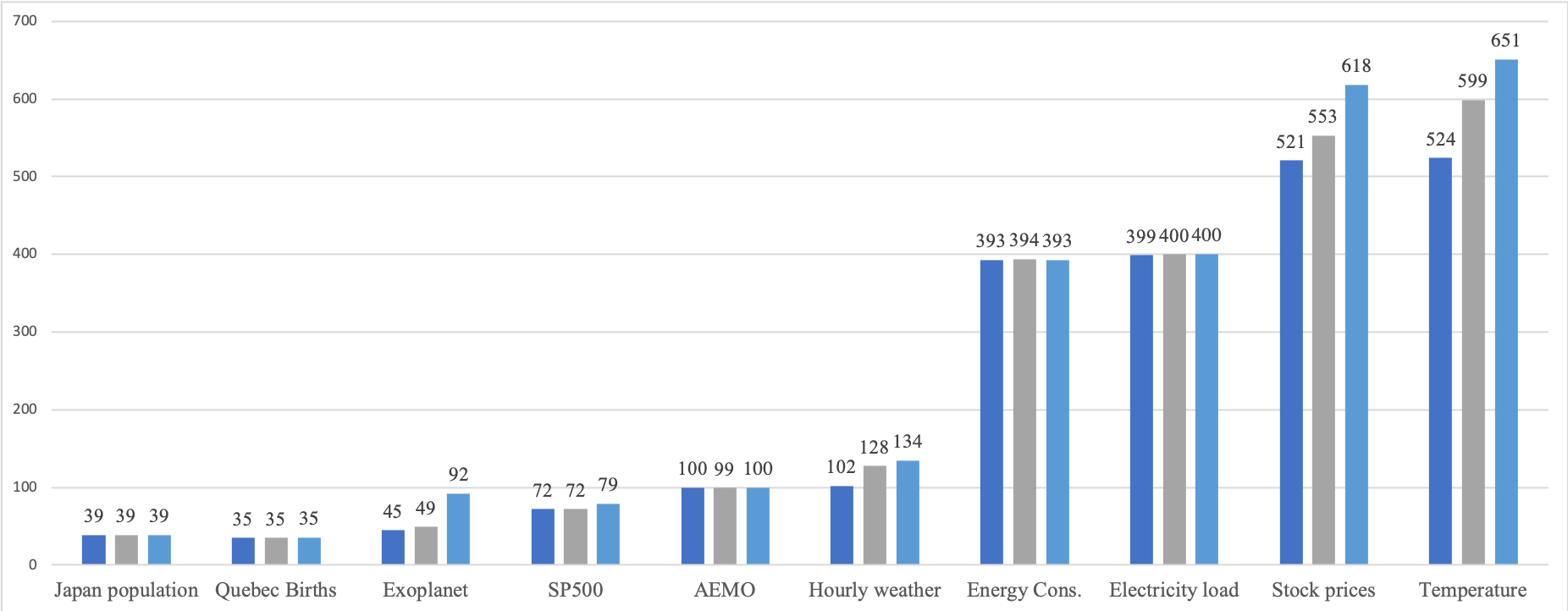}
		\caption{GRU}
		
	\end{subfigure}
	
	\caption{Speedup of \basic~and \opt~for the different architectures when $M=50$} 
	\label{fig:speedup}
\end{figure*}

Fig.~\ref{fig:speedup} illustrates the speedups of \basic~and \opt~for the six architectures tested when the number of hidden neurons $M$ is $50$. \opt~was tested with two different configurations: when the number of threads per block, block size BS, are $16$ and $32$, respectively. Clearly, \basic~and~\opt~achieve high speedups, especially when the size of the dataset increases. For instance, for the Elman architecture, \basic~achieves a speedup of $25$ on the small Exoplanet dataset, $99$ on the hourly energy consumption medium dataset and up to $399$ on the largest dataset (Temperature). However, \opt~does not always achieve higher speedups. Specifically, \basic~and \opt~achieve similar speedups for the Japan population, Quebec births, SP500, AEMO, energy consumption, and the electricity load datasets. 

To investigate these results, we take a closer look at the characteristics of the datasets. When $Q=10$,  a thread is computing the dot product between a row of $\bm{X}$ and a column of $\bm{W}$ and it is doing $2\times10$ memory read operations. Consequently, $num\_tiles$ will be only $1$ and the loop at line 8 of Alg.~\ref{algo:H_opt} will be only executed once. In this case, the performance does not improve and might slightly decrease due to the thread synchronization in \opt. However, \opt~achieves higher speedups when $Q>BS$ and when $BS$ increases to $32$. We notice that the speedup increases with more complex architectures, LSTM for example, since these architectures require more computations that can be better accelerated on a GPU. 

\subsection{Scalability}
\begin{figure*}
	\centering
	\begin{subfigure}[b]{0.95\textwidth}
		\centering
		\includegraphics[width=0.6\textwidth]{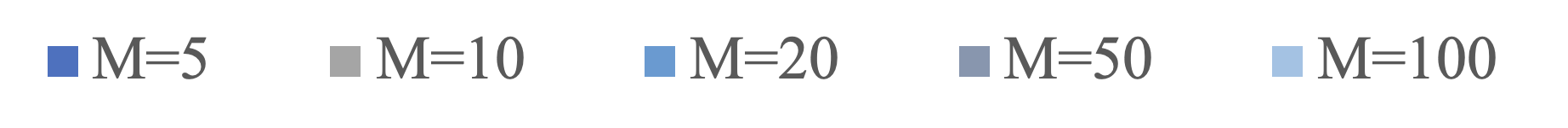}
	\end{subfigure}
	\begin{subfigure}[b]{0.475\textwidth}
		\centering
		\includegraphics[width=\textwidth]{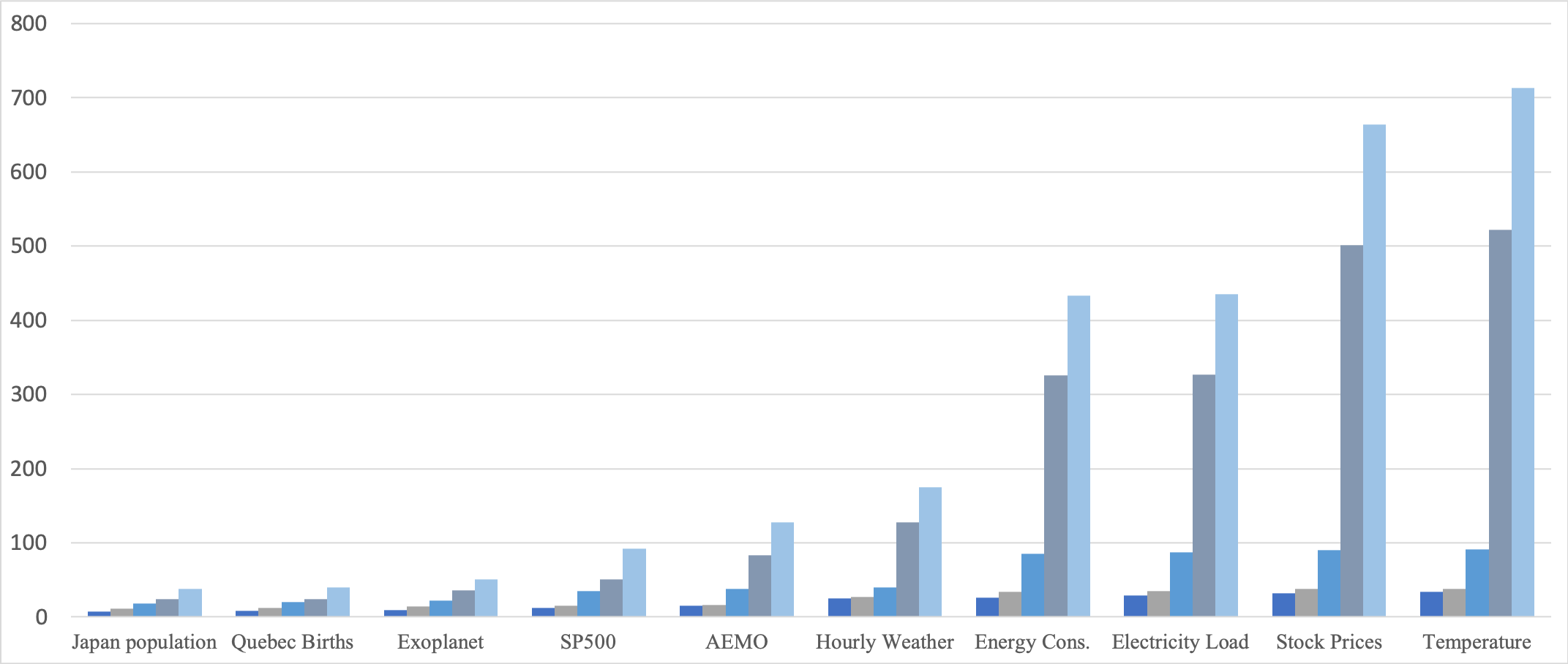}
		\caption{Jordan}
	\end{subfigure}
	\hfill
	\begin{subfigure}[b]{0.475\textwidth}  
		\centering 
		\includegraphics[width=\textwidth]{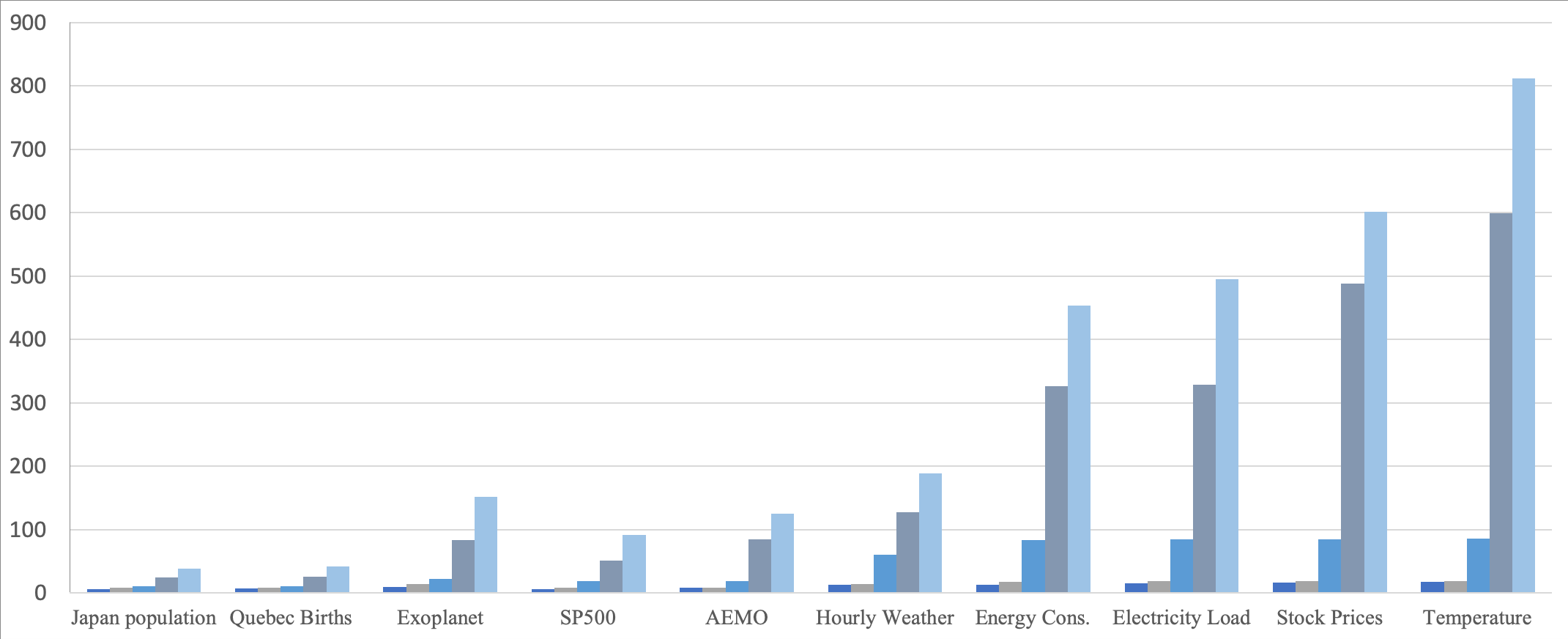}
		\caption{Elman}
	\end{subfigure}
	\vskip\baselineskip
	\begin{subfigure}[b]{0.475\textwidth}   
		\centering 
		\includegraphics[width=\textwidth]{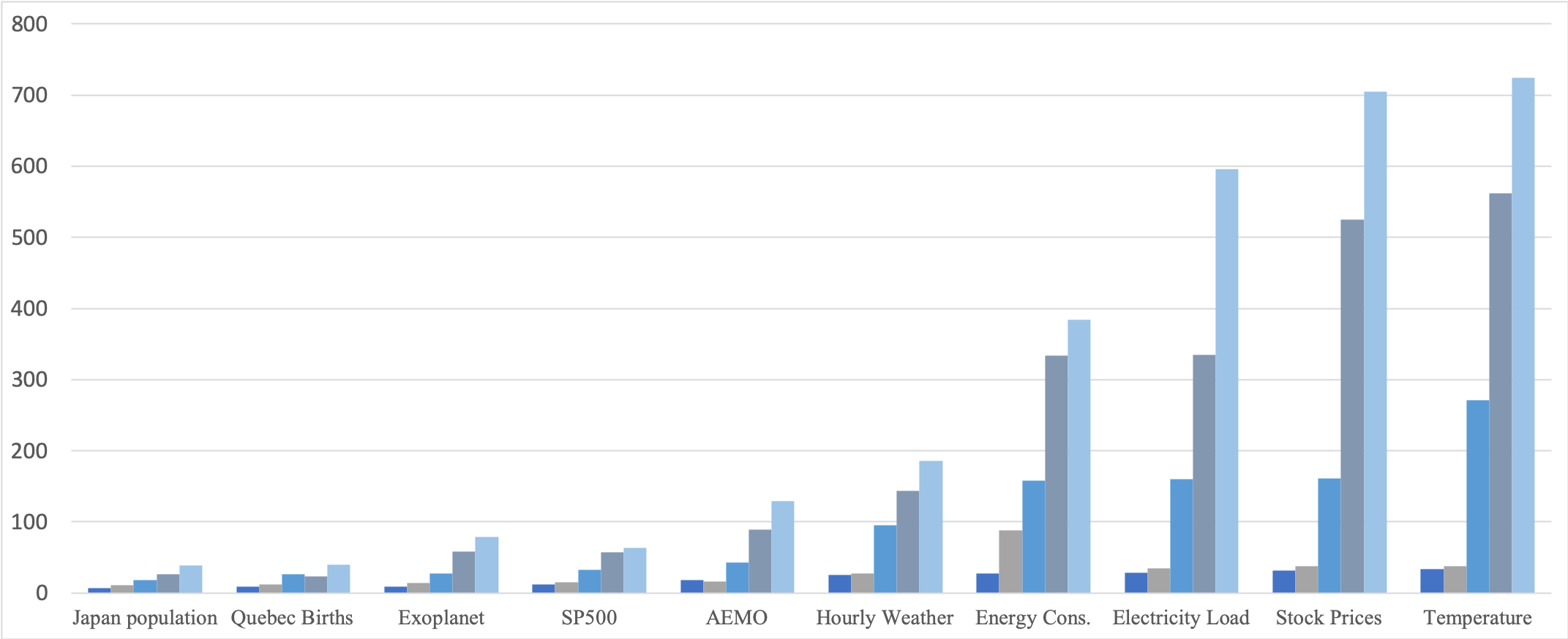}
		\caption{NARMAX}
	\end{subfigure}
	\quad
	\begin{subfigure}[b]{0.475\textwidth}   
		\centering 
		\includegraphics[width=\textwidth]{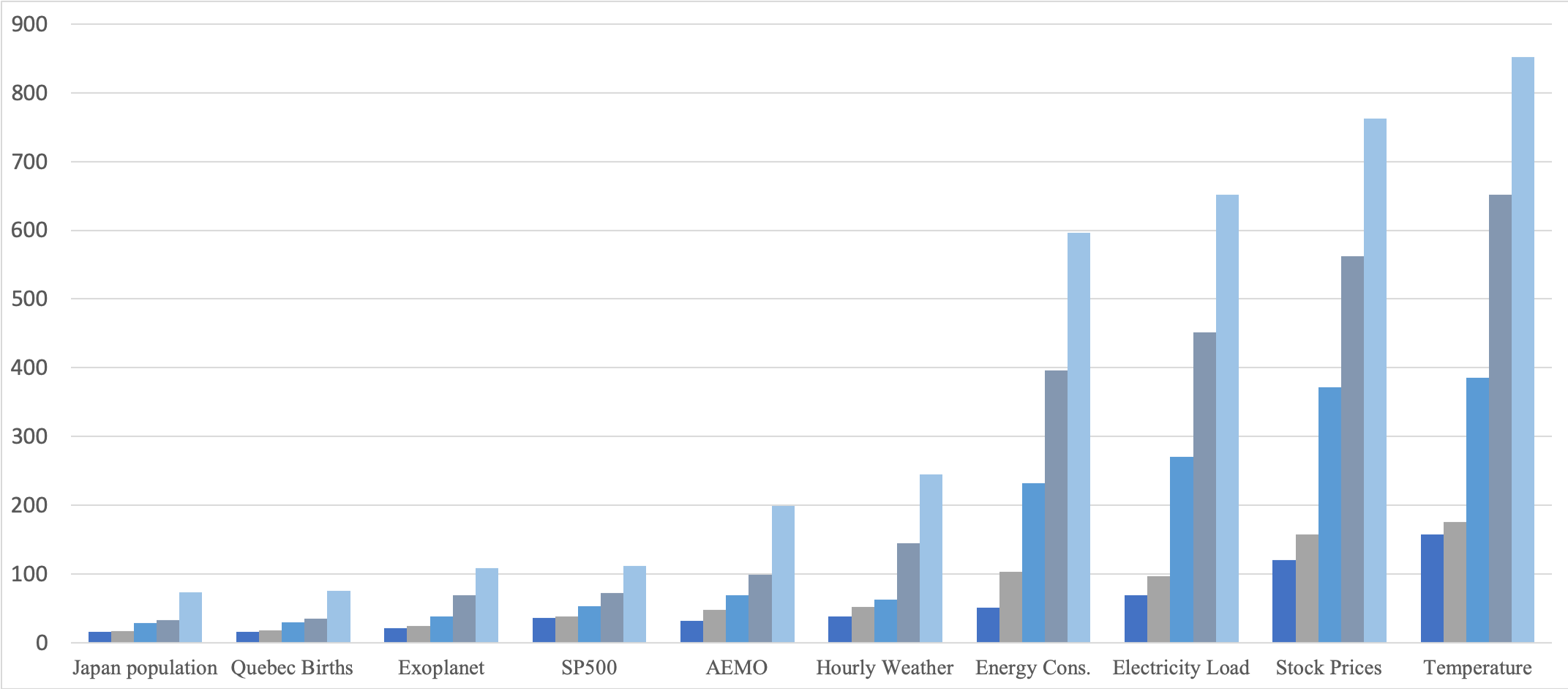}
		\caption{Fully Connected}
	\end{subfigure}
	\begin{subfigure}[b]{0.475\textwidth}   
		\centering 
		\includegraphics[width=\textwidth]{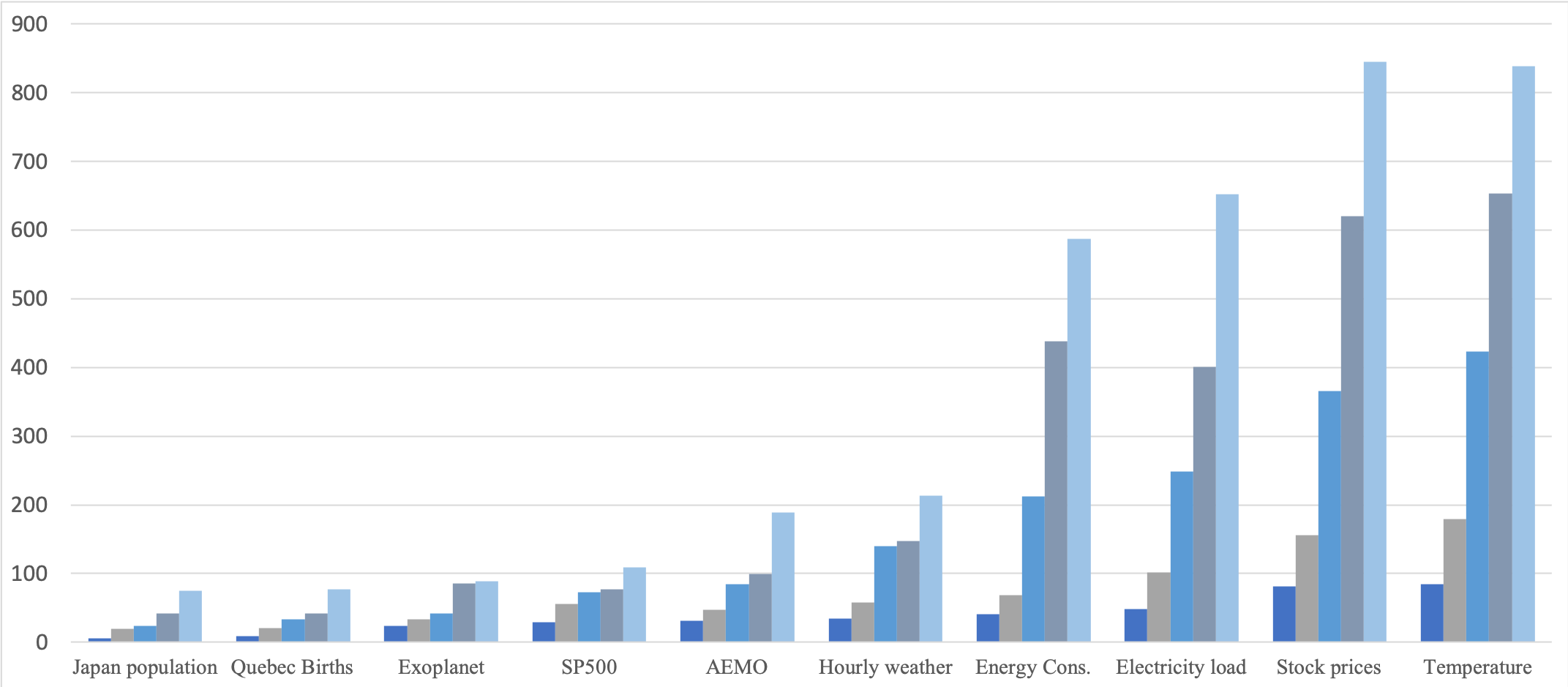}
		\caption{LSTM}
	\end{subfigure}
	\quad
	\begin{subfigure}[b]{0.475\textwidth}   
		\centering 
		\includegraphics[width=\textwidth]{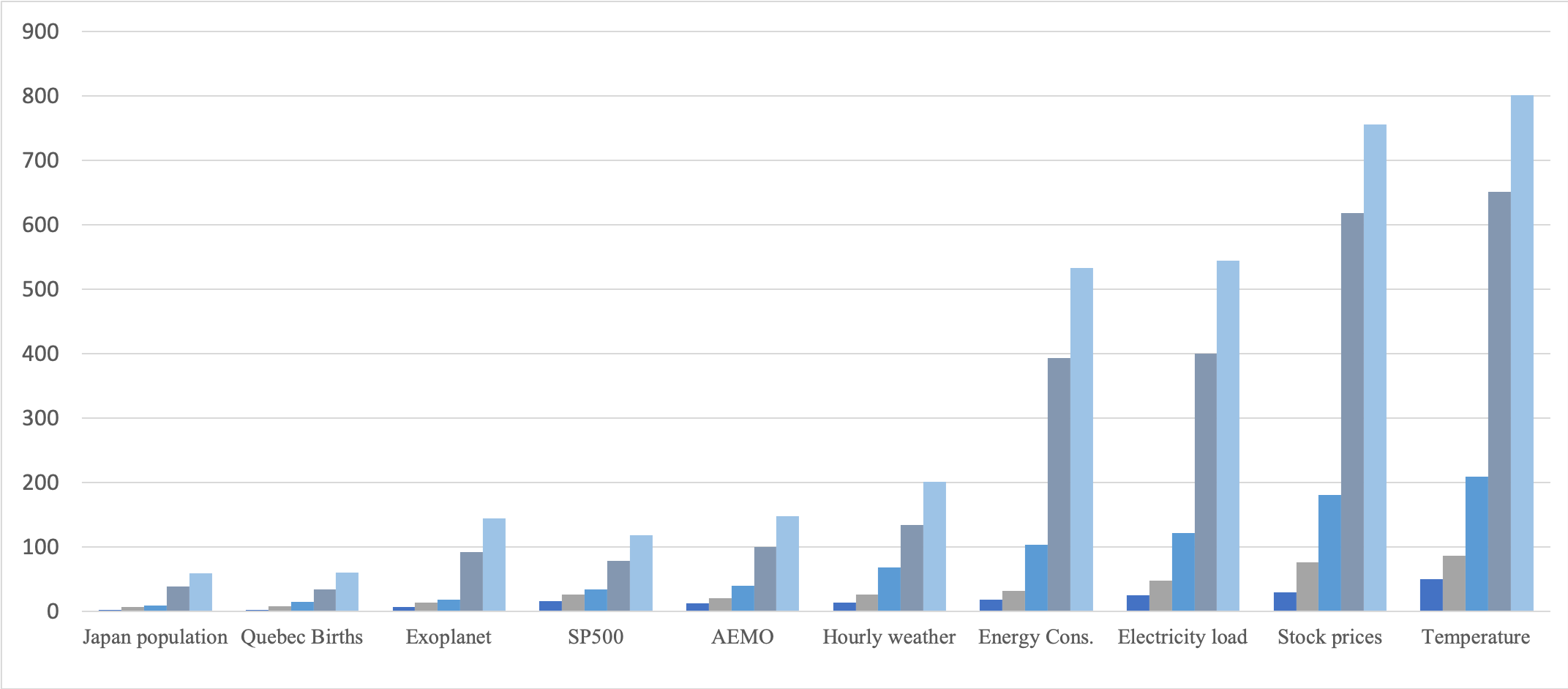}
		\caption{GRU}
		
	\end{subfigure}
	
	\caption{Speedup of \opt~for the different architectures when the number of hidden neurons increases from $5$ to $100$} 
	\label{fig:scal}
\end{figure*}

To test the scalability of our approach, we change the number of hidden neurons $M$, and we report the speedup of \opt~(BS=32) for the different architectures on the various datasets. Fig.~\ref{fig:scal} illustrates that the speedup increases $M$ increases from $5$ to $10,20,50,100$. Specifically, the speedup increases by a factor of 20 when $M$ increases from $5$ to $100$ with a GRU on the energy consumption dataset. Thus \opt~scales up well with more computationally expensive operations. 

\subsection{Robustness}
Robustness, i.e. repeatability, is a key property for \opt~where random initialization might affect the solution. Moreover, floating-point computations might differ between the GPU and the CPU, which might affect the output. To ensure that such perturbations do not affect the performance of our parallel algorithm, we run \seq~and \opt~(BS=32) five times, and we measure their root mean squared error (RMSE). Table~\ref{tbl:acc} reports the average RMSE and its standard deviation when \seq~and \opt~are tested on different datasets with different RNN architectures. We select $M$ according to the size of the problem; i.e. we used $M=100$ for exoplanet where $Q=5657$, $M=20$ for hourly weather, stock prices and temperature where $Q=50$ and $M=10$ for the rest of the datasets that have $Q=10$. Tables~\ref{tbl:db} and \ref{tbl:acc} show that the cases where the RMSE is high correspond to datasets with large outputs. For instance, having outputs ranging from $0$ to $2.06\times 10^9$, the electricity load dataset has higher RMSE than other datasets. However, \seq~and \opt~achieve accuracies in the same range for different RNN architectures on all the datasets, which means that GPU floating-point operations do not have a clear effect on the performance of our algorithm. 
\begin{sidewaystable*} 
	\caption{Average RMSE ($\pm$ standard deviation) of \seq~and \opt~(BS=32)}\label{tbl:acc}
	\begin{tabular}{p{1.1cm} l llllll}
		& & \multicolumn{6}{c}{\textbf{Architecture}} \\
		\cmidrule{3-8}
		\textbf{Dataset}&  \textbf{Algorithm}& \textbf{Elman} & \textbf{Jordan} & \textbf{NARMAX} & \textbf{Fully Connected} & \textbf{LSTM} & \textbf{GRU} \\
		\hline
		\hline
		\textbf{Japan pop.} & \textbf{\seq} & 3.97E-2 $\pm$ 4.67E-2 & 1.12E-1 $\pm$ 3.75E-1 & 6.54E-1 $\pm$ 3.32E-2 & 5.43E-3 $\pm$ 3.89E-5 & 2.45E-1 $\pm$ 2.36E-1 & 4.46E-1 $\pm$ 3.35E-4 \\
		& \textbf{\opt} & 3.74E-2 $\pm$ 7.17E-8 & 1.23E-1 $\pm$ 2.89E-2 & 6.23E-1 $\pm$ 2.31E-2 & 6.23E-3 $\pm$ 2.65E-4 & 2.46E-1 $\pm$ 4.56E-2 & 4.75E-2 $\pm$ 5.81E-5 \\
		\hline
		\textbf{Quebec Births} & \textbf{\seq} & 4.06E-3 $\pm$ 7.68E-5 & 1.01E-1  $\pm$  5.00E-3 & 3.42E-1  $\pm$  5.05E-3 & 2.02E-2  $\pm$  4.99E-7 & 1.01E-1 $\pm$ 5.76E-1 & 1.01E+0 $\pm$ 5.16E-4 \\
		& \textbf{\opt} & 2.02E-3 $\pm$ 4.89E-5 & 4.35E-1  $\pm$  5.32E-4 & 3.46E-1  $\pm$  3.79E-3 & 2.42E-2  $\pm$  7.07E-1 & 1.49E-2 $\pm$ 1.46E-4 & 1.16E+0 $\pm$ 3.56E-3 \\
		\hline
		\textbf{Exoplanet} & \textbf{\seq} & 5.40E+0 $\pm$ 3.03E-1 & 2.87E+0  $\pm$  7.91E-3 & 2.01E-1  $\pm$  2.98E-3 & 3.46E-1  $\pm$  1.01E-2 & 5.45E-1  $\pm$  2.31E-1 & 4.32E+0 $\pm$ 4.56E-1 \\
		& \textbf{\opt} & 5.42E+0 $\pm$ 3.05E-1 & 2.34E+0  $\pm$  7.34E-2 & 2.53E-1  $\pm$  1.98E-3 & 3.42E-1  $\pm$  1.51E-2 & 3.65E-1  $\pm$  2.31E-5 & 5.21E+0 $\pm$ 3.76E-2 \\
		\hline
		\textbf{SP500} & \textbf{\seq} & 1.69E-1 $\pm$ 7.78E-3 & 1.32E-1 $\pm$  3.75E-4 & 9.01E-1 $\pm$ 8.70E-4 & 1.96E+0 $\pm$  4.32E-1 & 1.01E-1 $\pm$ 5.16E-2 & 7.84E+0 $\pm$ 5.55E-2 \\
		& \textbf{\opt} & 2.34E-1 $\pm$ 7.98E-4 & 4.01E-1 $\pm$  6.36E-5 & 9.11E-1 $\pm$ 8.32E-5 & 1.36E+0 $\pm$  1.90E-2 & 1.24E-1 $\pm$ 3.14E-2 & 7.83E+0 $\pm$ 5.53E-1 \\
		\hline
		\textbf{AEMO} & \textbf{\seq} & 1.26E-1 $\pm$ 1.45E-3 & 3.30E-2 $\pm$ 7.16E-3 & 9.61E-2 $\pm$ 8.79E-3 & 5.00E-2 $\pm$ 1.32E-5 & 1.36E-2 $\pm$ 5.33E-4 & 2.33E-1 $\pm$ 2.23E-5 \\
		& \textbf{\opt} & 1.34E-1 $\pm$ 1.25E-4 & 1.12E-2 $\pm$ 5.16E-2 & 3.23E-3 $\pm$ 1.01E-2 & 5.36E-2 $\pm$ 1.12E-4 & 1.22E-2 $\pm$ 5.67E-3 & 2.01E-1 $\pm$ 2.13E-6 \\
		\hline
		\textbf{Hourly Weather} & \textbf{\seq} & 1.98E-1 $\pm$ 5.17E+0 & 3.14E-1 $\pm$ 2.07E-3 & 8.06E-1 $\pm$ 7.63E-5 & 7.39E-2 $\pm$ 6.03E-2 & 2.10E-2 $\pm$ 2.24E-5 & 3.21E-1 $\pm$ 9.61E-3 \\
		& \textbf{\opt} & 1.52E-1 $\pm$ 3.34E+0 & 3.98E-1 $\pm$ 5.67E-4 & 2.00E-1 $\pm$ 7.03E-4 & 3.79E-2 $\pm$ 5.03E-3 & 1.02E-2 $\pm$ 2.14E-5 & 4.32E-1 $\pm$ 9.16E-3 \\
		\hline
		\textbf{Energy Cons.} & \textbf{\seq} & 1.83E-4 $\pm$ 1.98E-3 & 2.21E-3 $\pm$ 3.43E-1 & 2.22E-4 $\pm$ 5.26E-3 & 3.56E-3 $\pm$ 5.56E-4 & 1.56E-3 $\pm$ 9.96E-4 & 2.34E-2 $\pm$ 2.22E-5 \\
		& \textbf{\opt} & 1.38E-4 $\pm$ 2.45E-3 & 3.48E-3 $\pm$ 3.03E-2 & 6.44E-5 $\pm$ 5.16E-4 & 2.65E-3 $\pm$ 5.16E-5 & 2.56E-3 $\pm$ 5.326E-5 & 3.24E-3 $\pm$ 2.12E-5 \\
		\hline
		\textbf{Elec. Load }& \textbf{\seq} & 2.56E+0 $\pm$ 7.93E+0 & 2.40E+0 $\pm$ 3.90E-1 & 8.64E+0 $\pm$ 9.81E+0 & 4.16E-1 $\pm$ 3.45E-1 & 8.32E+0 $\pm$ 8.05E+0 & 1.12E+0 $\pm$ 5.16E-1 \\
		& \textbf{\opt} & 2.34E+0 $\pm$ 7.03E-1 & 4.76E+0 $\pm$ 2.20E-2 & 4.86E+0 $\pm$ 8.91E-1 & 4.64E-1 $\pm$ 3.97E-2 & 2.84E+0 $\pm$ 8.13E-1 & 2.98E+0 $\pm$ 5.06E+0 \\
		\hline
		\textbf{Stock Prices }& \textbf{\seq} & 6.41E-1 $\pm$ 7.93E-1 & 1.10E-1 $\pm$ 9.09E-5 & 4.80E+0 $\pm$ 3.87E-1 & 2.13E-2 $\pm$ 3.89E-1 & 4.00E-1 $\pm$ 1.09E-3 & 2.62E-1 $\pm$ 3.82E-4 \\
		& \textbf{\opt} & 3.41E-1 $\pm$ 3.35E-2 & 1.56E-1 $\pm$ 9.23E-5 & 4.81E+0 $\pm$ 3.32E-2 & 2.03E-3 $\pm$ 1.92E-4 & 4.94E-1 $\pm$ 5.69E-4 & 6.28E-1 $\pm$ 3.28E-3 \\
		\hline
		\textbf{Temp.} & \textbf{\seq} & 4.32E-4 $\pm$ 9.85E-5 & 5.65E-3 $\pm$ 6.79E-9 & 3.56E-4 $\pm$ 7.10E-6 & 2.91E-5 $\pm$ 3.72E-9 & 4.92E-4 $\pm$ 6.02E-5 & 3.54E-4 $\pm$ 2.95E-6 \\
		&  \textbf{\opt} & 4.12E-4 $\pm$ 9.67E-4 & 5.03E-3 $\pm$ 6.19E-2 & 3.15E-4 $\pm$ 9.25E-6 & 9.21E-5 $\pm$ 3.02E-5 & 8.17E-4 $\pm$ 6.92E-4 & 3.19E-3 $\pm$ 5.29E-5 \\
		\hline
	\end{tabular}
	
\end{sidewaystable*} 

\subsection{Portability}
To verify that our algorithm is portable, we ran \opt~(BS=32) on an NVIDIA Quadro K2000 GPU while fixing the number of hidden nodes $M$ at $50$. It is important to check for portability to understand how much the proposed algorithm is architecture dependent. Table~\ref{tbl:porta} shows that \opt~also achieves high speedups on the Quadro K2000 GPUs for different RNN architectures on different datasets but the speedups on the Tesla K20m GPU are constantly higher because of the computational capability of the latter. 
\begin{table*}[]
	\caption{Speedup of \opt~(BS=32) when tested on the Tesla K20m and Quadro K2000 GPUS for different RNN architectures on various datasets when the number of hidden neurons $M$ is $50$.}\label{tbl:porta}
	\begin{tabular}{ l l  p{1.1cm}p{1.1cm}p{1.1cm}p{1.1cm}p{1.1cm}p{1.1cm}p{1.1cm}p{1.1cm}p{1.1cm}p{1.1cm} }
		& & \multicolumn{10}{c}{\textbf{Dataset}} \\
		\cmidrule{3-12}
		\textbf{Architecture} & \textbf{GPU} & \textbf{Japan pop.} & \textbf{Quebec Births} & \textbf{Exoplanet} & \textbf{SP500} & \textbf{AEMO} & \textbf{Hourly weather} & \textbf{Energy cons.} & \textbf{Elec. Load }& \textbf{Stock Prices} & \textbf{Temp.} \\
		\hline
		\hline
		\textbf{Elman} & \textbf{Tesla}  & 24 & 24 & 36 & 51 & 83 & 128 & 326 & 327 & 501 & 522 \\
		& \textbf{Quadro}  & 23 & 19 & 32 & 45 & 79 & 121 & 120 & 326 & 478 & 501 \\
		\hline
		\textbf{Jordan} & \textbf{Tesla}  & 24 & 25 & 83 & 51 & 84 & 127 & 326 & 329 & 488 & 599 \\
		& \textbf{Quadro}  & 22 & 21 & 78 & 45 & 79 & 120 & 325 & 325 & 378 & 589 \\
		\hline
		\textbf{NARMAX} & \textbf{Tesla}  & 26 & 23 & 58 & 57 & 89 & 144 & 334 & 335 & 525 & 562 \\
		& \textbf{Quadro}  & 21 & 21 & 55 & 52 & 83 & 141 & 323 & 324 & 513 & 545 \\
		\hline
		\textbf{Fully } & \textbf{Tesla}  & 33 & 35 & 69 & 72 & 99 & 145 & 396 & 452 & 562 & 652 \\
		\textbf{Connected}& \textbf{Quadro}  & 28 & 32 & 65 & 67 & 96 & 141 & 391 & 449 & 558 & 647 \\
		\hline
		\textbf{LSTM} & \textbf{Tesla}  & 42 & 42 & 86 & 77 & 99 & 147 & 438 & 401 & 620 & 653 \\
		& \textbf{Quadro}  & 37 & 39 & 82 & 72 & 89 & 140 & 430 & 391 & 613 & 645 \\
		\hline
		\textbf{GRU} & \textbf{Tesla}  & 39 & 35 & 92 & 79 & 100 & 134 & 393 & 400 & 618 & 651 \\
		& \textbf{Quadro}  & 29 & 28 & 84 & 69 & 93 & 116 & 383 & 373 & 600 & 639\\
		\hline
	\end{tabular}
	
\end{table*}
\subsection{Energy Efficiency}
Recently, designing less power-consuming models is becoming of great importance when high-performance workstations are implemented. Alongside speedup, we consider power consumption as an essential metric according to which \opt~is evaluated. For instance, based on past experience, the CPU used in the benchmarks uses at least $30$ Watts when performing heavy computations (such as Moore-Penrose Pseudo inverse), whereas the GPU uses around $300$ Watts. Hence, whenever \basic~or \opt~exhibit a speedup higher than $10$, they not only become faster than \seq~but more power-efficient. In particular, \opt~needs $3.71$ seconds, consuming $1,113$ Joules, to train an RNN with Elman architecture and $M=50$ , whereas \seq~needs $32$ minutes to complete the same task. Thus, for this configuration, \seq~consumes $57,600$ Joules on the CPU, i.e. 50x more energy than \opt~ on the GPU. Such results are considerably important in time-series prediction applications with power constraints. For instance, this GPU implementation comes in handy in seismic monitoring where power might be provided by solar devices, and online computations are performed every few milliseconds.

\subsection{Comparison with Parallel Iterative RNN Training}
\begin{figure}
	\centering
	\includegraphics[width=0.49\textwidth]{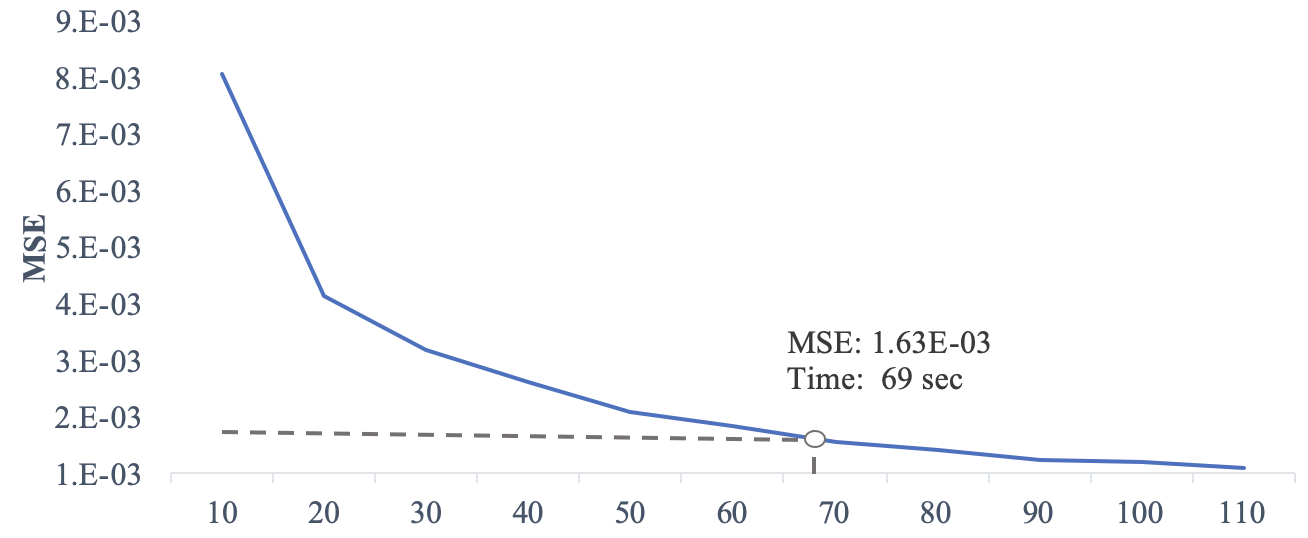}
	\caption{MSE versus time (sec) for P-BPTT algorithms when tested on the Japan population dataset with $M=10$ and LSTM as architecture}\label{fig:vs_it}
\end{figure}

Although \opt~achieves high speedups compared its \seq, we need to show that its absolute training time is lower than the parallel version of the BPTT (P-BPTT) as implemented in \cite{tensorflow2015-whitepaper}. We choose the architectures that \cite{tensorflow2015-whitepaper} implements, i.e. fully connected, LSTM and GRU, and we report the training time of \opt~(BS=32) and P-BPTT when $M=10$. P-BPTT is trained for $10$ epochs with $64$ as batch size, \textit{mean squared error} (MSE) as loss function and \textit{ADAM} as optimizer. We are interested in the absolute training times of the two parallel algorithms rather than their speedup over their sequential versions. Thus, we report the runtimes of \opt~and P-BPTT algorithms when tested on the same Tesla K20m GPU and the ratio between both training times. As Table~\ref{tbl:vs_it} shows, \opt~runs up to $20$x faster than P-BPTT when tested with LSTM on the Japan population dataset.\\
Fig.~\ref{fig:vs_it} illustrates the MSE versus time for P-BPTT algorithms when tested with LSTM on the Japan population dataset with $M=10$. For the same dataset and RNN architecture, \opt~reaches $1.63\times 10^{-3}$ as MSE, whereas P-BPTT reaches a lower MSE of $1.1\times 10^{-3}$. However, \opt~took only $0.07$ sec to reach its optimal MSE, whereas P-BPTT took $110$ sec to reach its optimal MSE and $69$ sec to reach the same MSE ($1.1\times 10^{-3}$). Thus, \opt~could reach the same performance as P-BPTT $956$x slower. The sequential nature of iterative training explains the results: although one can attempt to parallelize each epoch, the training needs to be done in a sequence of consecutive dependent epochs. 

\begin{table*}[]
	\caption{Runtime (seconds) of \opt~(BS=32) and the iterative training algorithm and the ratio=$\frac{\text{BP}}{\text{\opt}}$}\label{tbl:vs_it} 
	\begin{tabular}{l r r r r r r r r r }
		& \multicolumn{3}{c}{\textbf{Fully Connected}} & \multicolumn{3}{c}{\textbf{LSTM}} & \multicolumn{3}{c}{\textbf{GRU}} \\
		\cmidrule(l{5pt}r{2pt}){2-4} \cmidrule(l{5pt}r{2pt}){5-7} \cmidrule(l{5pt}r{2pt}){8-10}
		
		& \textbf{Opt-PR-ELM} & \textbf{P-BPTT} &  \textbf{ Ratio} & \textbf{Opt-PR-ELM} & \textbf{P-BPTT} &  \textbf{Ratio} & \textbf{Opt-PR-ELM} & \textbf{P-BPTT} &  \textbf{Ratio} \\
		\hline\hline
		\textbf{Japan pop. }& 0.23 & 3.52 & 15 & 0.38 & 7.41 & 20 & 0.38 & 6.59 & 17 \\
		\textbf{Quebec Births }& 0.56 & 6.75 & 12 & 0.85 & 13.56 & 16 & 0.81 & 12.94 & 16 \\
		\textbf{Exoplanet} & 10.03 & 24.98 & 2 & 15.23 & 54.32 & 4 & 13.14 & 43.12 & 3 \\
		\textbf{SP500} & 3.56 & 20.66 & 6 & 7.77 & 37.55 & 5 & 5.61 & 35.65 & 6 \\
		\textbf{AEMO} & 3.01 & 21.34 & 7 & 7.29 & 38.32 & 5 & 5.62 & 35.71 & 6 \\
		\textbf{Hourly Weather} & 30.46 & 156.76 & 5 & 50.49 & 243.99 & 5 & 30.04 & 201.12 & 7 \\
		\textbf{Energy Cons. }& 32.14 & 203.45 & 6 & 51.90 & 525.87 & 10 & 45.67 & 435.89 & 10 \\
		\textbf{Elec. Load} & 36.70 & 256.89 & 7 & 53.60 & 572.74 & 11 & 51.7 & 532.31 & 10 \\
		\textbf{Stock Prices} & 41.30 & 301.23 & 7 & 56.78 & 639.04 & 11 & 52.34 & 621.18 & 12 \\
		\textbf{Temperature} & 45.45 & 354.99 & 8 & 62.00 & 678.11 & 11 & 59.32 & 641.09 & 11\\
		\hline
	\end{tabular}
	
\end{table*}

\subsection{\opt ~Run Time}
\begin{figure}
	\includegraphics[width=0.49\textwidth]{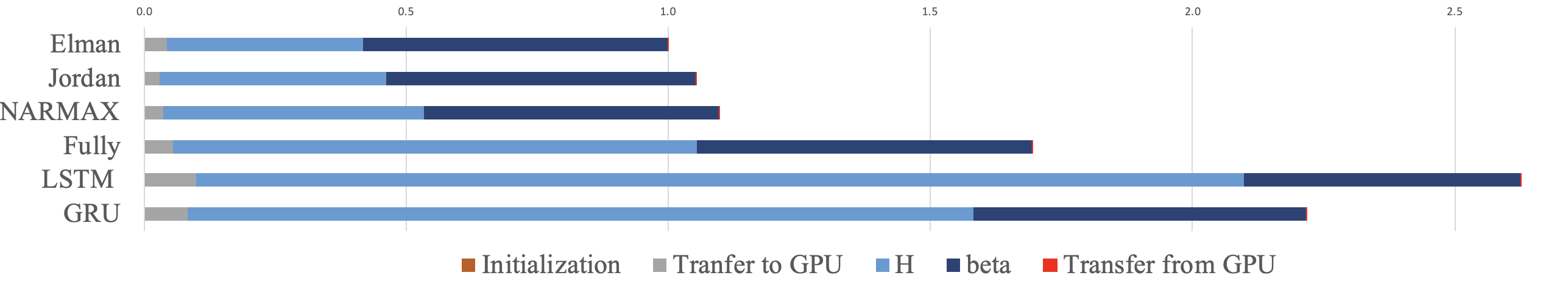}
	\caption{Time decomposition (sec) of \opt~on the Japan population dataset with $M=10$}\label{fig:time_decom}
\end{figure}

One can argue that using memory streams or initializing the random weights on the GPU can lead to higher speedups. To investigate this, we study how the runtime of \opt~ is decomposed between the parameters initialization, data transfer to and from the GPU and the actual computations for the six architectures. Fig.~\ref{fig:time_decom} shows what portion each step takes from the runtime of \opt~when tested on the Japan population dataset with $M=10$. The initialization does not appear on the bar because it is less than $0.01\%$ of the total runtime. Moreover, transfer data to the GPU consistently takes more time than the transfer back because the former deals with the following matrices: $\bm{X} \in \mathbb{R}^{n \times S \times Q}$, $\bm{Y} \in \mathbb{R}^{n}$, $\bm{W} \in \mathbb{R}^{S \times L}$, $\bm{\alpha} \in \mathbb{R}^{L \times Q} $ and $\bm{b} \in \mathbb{R}^{L}$, while the latter only transfers $\beta \in \mathbb{R}^{L}$. The steps that take the major time portion are the computations of $\bm{H}$ and $\beta$. One can conclude that data streams or the GPU random initializations will not affect the speedup since initialization and data transfer are not a bottleneck in \opt. 

\section{Conclusion}\label{sec:conc}
In this work, we proposed \opt, a parallel version of non-iteratively trained RNNs for time series prediction. Focusing on six RNN architectures: Elman, Jordan, NARMAX, fully connected RNN, LSTM and GRU, we first developed a basic version of the parallel algorithm and. Then, we studied its memory access patterns to propose an optimized version that takes advantage of the shared memory of the GPU. In addition to performing a theoretical, computational analysis of \opt~on the various architectures, empirical validation was performed on $10$ publicly available time series prediction datasets. 

\opt~was shown to achieve a speedup of up to $845$ over its sequential version and up to $20$ over the parallel BPTT version. We further studied the scalability of our proposed algorithm by changing the number of hidden neurons and reporting the speedup. \opt~ showed higher speedups when the number of computations increases or the number of launched threads per block increases.  Moreover, portability of \opt~was studied when it is tested on a different GPU architecture where it reached a high speedup of up to $647$. Finally, \opt~was shown to reach similar accuracies as its sequential version while consuming less energy.  

Future work includes extending \opt~to RNNs with multiple layers and investigating its performance on applications that have multi-dimensional outputs such as machine translation and speech recognition.


%



\section*{Acknowledgment}
This work was supported by the University Research Board at the American University of Beirut.
\ifCLASSOPTIONcaptionsoff
  \newpage
\fi

\bibliographystyle{plain}
\bibliography{cit}

\vspace{-3em}
\begin{IEEEbiography}[{\includegraphics[width=1in,height=1.25in,clip,keepaspectratio]{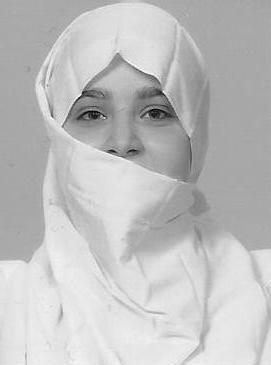}}]{Julia El Zini}
	is a PhD student enrolled in the electrical and computer engineering department at the American University of Beirut (AUB). She has received her BS and MS in computer science from AUB, Lebanon, in 2015 and 2017, respectively.\\
	Her research interests include distributed optimization, parallel computing, reinforcement leaning, multi-task and transfer learning, and scalable machine learning applications.
\end{IEEEbiography}
\vspace{-3.5em}
\begin{IEEEbiography}[{\includegraphics[width=1in,height=1.25in,clip,keepaspectratio]{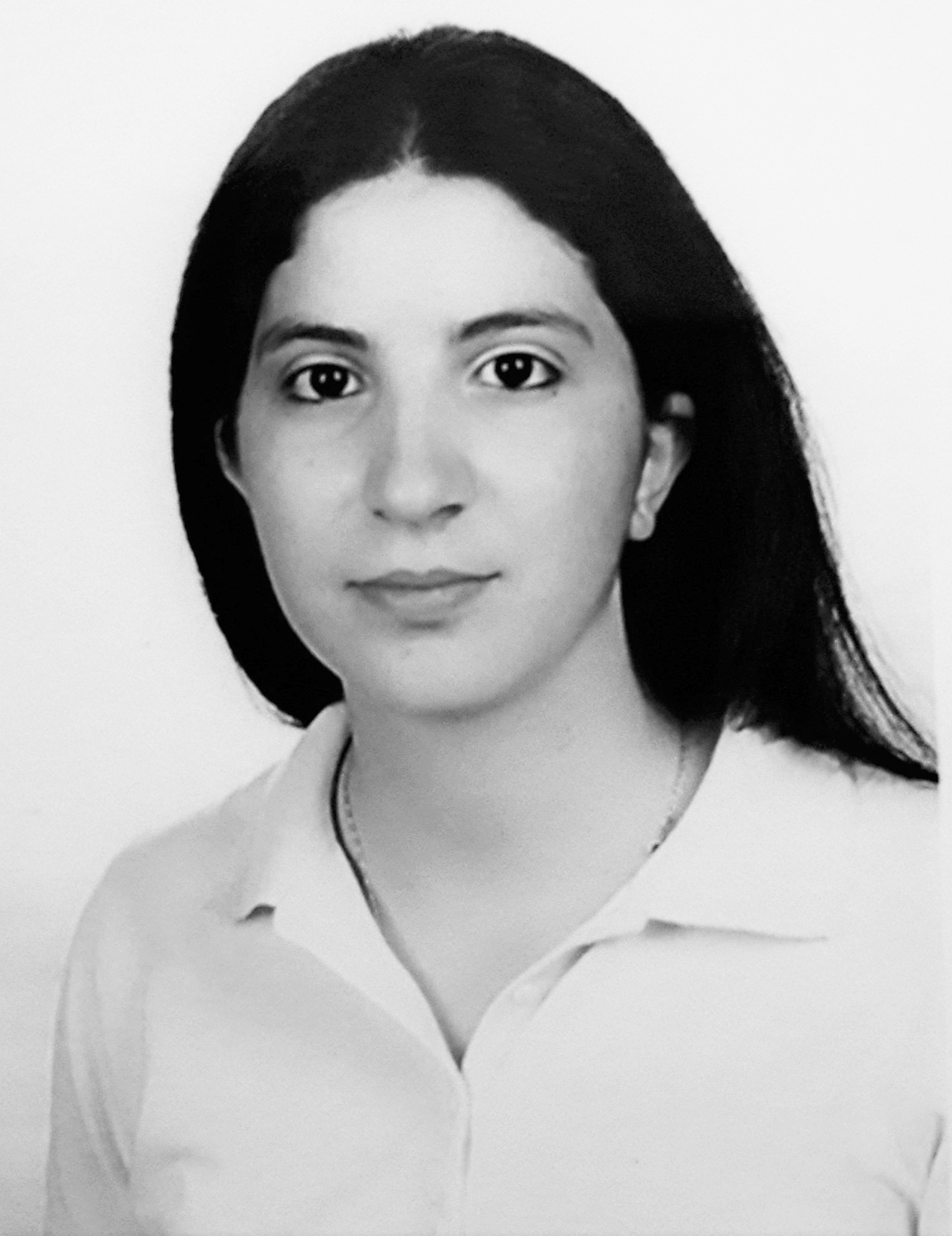}}]{Yara Rizk}
	obtained her PhD in Electrical and Computer Engineering from the American University of Beirut (AUB) in 2018. Prior, she received her BE in Computer and Communication Engineering from AUB, Lebanon, in 2012. 
	Her research interests span robotics, multi-agent systems, machine learning, classification, clustering, and artificial intelligence.
	Rizk has attended a technical internship (2013-2014) at Intel in Hillsboro, Oregon, USA and is an active researcher multiple peer-reviewed publications. 
\end{IEEEbiography}

%

\vspace{-3.5em}
\begin{IEEEbiography}[{\includegraphics[width=1in,height=1.25in,clip,keepaspectratio]{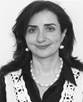}}]{Mariette Awad}
	(M'08) obtained her PhD in Electrical Engineering from the University of Vermont (2007). Her current research focuses on HMI, efficient artificial intelligence, applied machine learning and Internet of Things. Dr. Awad has received more than 25 grants to support her research including 2 multidisciplinary multi-million dollar grants from the Qatar National Research Fund (QNRF) and Intel. Her work culminated in a book, Efficient Machine Learning, in 2015 as well as more than 100 conference, book chapter, and journal publications. Prior to her academic position, she was with IBM System and Technology group in Vermont for six years where her technical leadership and innovative spirit has earned her management recognition twice, two business awards, and 10 patents.
\end{IEEEbiography}




\end{document}